\documentclass{article}

\PassOptionsToPackage{numbers,sort,compress}{natbib}
\usepackage[preprint]{neurips_2026}

\usepackage[utf8]{inputenc} 
\usepackage[T1]{fontenc}    
\usepackage{hyperref}       
\usepackage{url}            
\usepackage{booktabs}       
\usepackage{amsfonts}       
\usepackage{nicefrac}       
\usepackage{microtype}      
\usepackage[dvipsnames]{xcolor}         

\usepackage{graphicx}
\usepackage{amsmath}
\usepackage{subcaption}
\usepackage{titletoc}

\newcommand{\E}{\mathcal{E}}
\newcommand{\D}{\mathcal{D}}

\usepackage{tabularx}
\usepackage[table]{xcolor}
\usepackage{multirow}
\usepackage{wrapfig}

\title{FLUXtrapolation: \\ A benchmark on extrapolating ecosystem fluxes}

\author{%
    Anya Fries \\
    Seminar for Statistics, ETH Z\"urich\\
    Z\"urich, Switzerland \\
    \texttt{anya.fries@stat.math.ethz.ch} \\
    \And
    Jacob A. Nelson \\
    Max Planck Institute for Biogeochemistry \\
    Jena, Germany \\
    \texttt{jnelson@bgc-jena.mpg.de} \\
    \And
    Martin Jung \\
    Max Planck Institute for Biogeochemistry \\
    Jena, Germany \\
    \texttt{mjung@bgc-jena.mpg.de} \\
    \And
    Markus Reichstein \\
    Max Planck Institute for Biogeochemistry \\
    Jena, Germany \\
    \texttt{mreichstein@bgc-jena.mpg.de} \\
    \And
    Jonas Peters \\
    Seminar for Statistics, ETH Z\"urich\\
    Z\"urich, Switzerland \\
    \texttt{jonas.peters@stat.math.ethz.ch} \\
}

\begin{document}

\maketitle

\begin{abstract}
We introduce FLUXtrapolation, a benchmark for extrapolating ecosystem fluxes under progressively harder distribution shifts. Ecosystem fluxes are central to understanding the carbon, water, and energy cycles, yet they can only be measured directly at sparsely located measurement towers. Producing global flux estimates therefore requires training models on observed sites using globally available covariates and predicting in unobserved regions, that is, upscaling. Flux upscaling is a challenging domain generalization problem that is affected by a shift in covariate distribution across climates, ecosystem types, and environmental conditions, as well as by conditional shift: important drivers remain unobserved at global scale. We provide a quantitative analysis of both these shifts in $P_X$ and $P_{Y\mid X}$. FLUXtrapolation is designed based on domain expertise on flux upscaling: it defines temporal, spatial, and temperature-based extrapolation scenarios and evaluates performance across held-out domains, temporal aggregations, and tail errors. In a pilot study, we find that baselines perform similarly under median hourly RMSE, but separate under the proposed tail-focused and multi-scale evaluation. FLUXtrapolation therefore poses a realistic and thus relevant challenge for machine learning methods under distribution shift; at the same time, progress on this benchmark would directly support the scientific goal of improving flux upscaling. 
\end{abstract}

\section{Introduction}\label{sec:intro}
\begin{figure}[t]
    \centering
    \includegraphics[width=\linewidth, trim={0 24.5cm 0 0}, clip]{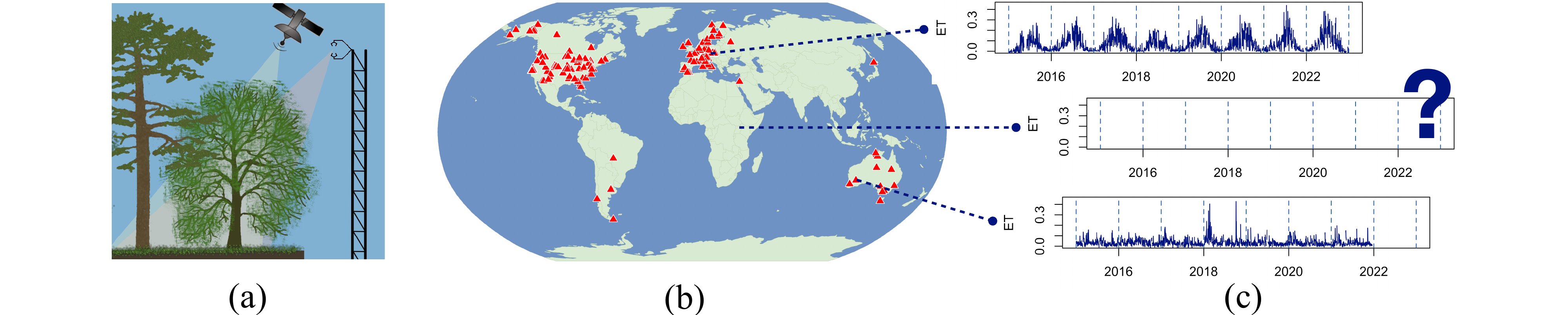}
    \caption{
    Flux tower data and the upscaling challenge.
    (a)~An eddy covariance tower, which measures ecosystem–atmosphere exchanges (such as carbon, water, energy).     
    (b)~Tower locations are sparsely located,
    motivating spatial upscaling. 
    (c)~Data at two 
    locations with towers (the figure shows daily averages; the benchmark uses hourly data). The ultimate goal is to predict at towerless locations. FLUXtrapolation approximates this problem by holding out tower sites or site-years.
    }
    \label{fig:overall}
\end{figure}

The exchange of carbon, water, and energy between terrestrial ecosystems and the atmosphere are central to understanding and modeling the Earth system. 
Direct observations of these exchanges, or \emph{ecosystem fluxes}, are available from eddy-covariance towers and compiled at scale in networks such as FLUXNET \citep{pastorello2020fluxnet2015}.
As
these towers are sparsely located globally,
producing global flux estimates
requires predicting fluxes at locations without towers, a task known as \emph{upscaling} 
(see Figure~\ref{fig:overall}). 
Upscaling is a challenging domain generalization and extrapolation problem:
towerless locations can differ substantially from observed tower sites in 
terms of
climate, ecosystem characteristics, and other environmental conditions.
Moreover, 
important drivers of fluxes 
such as soil and understory properties are
unobserved, 
effectively yielding shifts in the conditional distribution.

Flux upscaling poses a form of domain generalization that is not well reflected in existing benchmarks (detailed in Section~\ref{sec:related-work}). It consists of spatiotemporal data, many heterogeneous domains, and shifts both in the inputs and in the input--target relationship. In this setting, average test error alone can mask substantial failures on particular sites or site-years and says little about whether a model remains accurate across the several temporal scales relevant for scientific use.
Accordingly, 
our evaluation 
assesses performance 
beyond mean and median and
across different 
domains 
and temporal scales.

When predicting at
new locations, 
extrapolation
need not arise in the same way everywhere. In some regions, the main difference lies in the observed covariates; in others, unobserved biological drivers can change the relationship between available inputs and ecosystem fluxes. Making this shift structure explicit is therefore important both for interpreting benchmark results \citep{language2023liu, cai2025diagnosing} and informs our design of evaluation settings with progressively harder extrapolation.

We introduce FLUXtrapolation, a benchmark for ecosystem flux prediction under progressively harder 
extrapolation. It defines 
extrapolation scenarios that vary in difficulty and 
in the type of distribution shift they induce.
Drawing on state-of-the-art flux upscaling evaluation
\citep{tramontana2016, xbase2024, scalingcarbon2020jung}, 
our evaluation protocol assesses robustness across domains and 
multiple temporal scales. 
FLUXtrapolation 
is a realistic distribution-shift challenge for machine learning 
and,
since it builds on domain science expertise, any progress 
may generate direct impact in the field of Earth system science; lastly, it may 
also inform
other
upscaling problems such as 
for forest and soil properties \citep{forestproperties, soilproperties}.

\subsection{Related work}\label{sec:related-work}

\paragraph{Existing domain generalization benchmarks in ML.} 
Existing domain generalization benchmarks are 
often
dominated by image-based settings \citep[e.g.,][]{gulrajani2021domainbed, koh2021wilds}. 
Recent work has also considered temporal shift, most notably Wild-Time \citep{yao2022wildtime}, and tabular data under shift, such as TableShift and WhyShift \citep{tableshift2023gardner, language2023liu}.
Flux upscaling is nevertheless distinct: it is a spatiotemporal prediction problem with many heterogeneous domains and structured extrapolation across sites and site-years. While WhyShift makes shift structure explicit, 
flux upscaling adds the 
requirement that the benchmark reflect the evaluation needs of the scientific task, rather than considering single aggregate metrics.

\paragraph{Evaluating performance under distribution shift.}
A single aggregate metric 
can be
insufficient 
under distribution shift, since degradation can arise from different sources
\citep{cai2025diagnosing}.
Aggregate metrics may 
also 
hide poor performance on rare but important subpopulations \citep{li2021worstcase}. 
In spatial and environmental prediction, pooled summaries can 
obscure substantial variation
so evaluation should reflect 
intended deployment 
\citep{meyer2022globalmaps, ploton2020spatial}. Our evaluation protocol adapts this perspective to 
flux prediction by 
assessing performance 
across heterogeneous 
domains and scientifically relevant temporal scales.

\paragraph{Flux upscaling and benchmarks.} 
In Earth system science, flux upscaling has been developed 
prominently through FLUXCOM (FLUX COMparison), 
a scientifically grounded framework for constructing and evaluating global flux products, with a focus on evaluating uncertainty, including  feature choice, model class, 
target uncertainty,
and extrapolation \citep{tramontana2016, scalingcarbon2020jung}. 
Its current iteration, FLUXCOM-X (FX), 
provides updated upscaling products \citep{xbase2024} and continued evaluation \citep{kraft2025sequential}. 
X-BASE, the baseline FX product configuration, is built using XGBoost, a strong-performing standard in this setting. 
FLUXtrapolation adapts
FX into a 
domain generalization
benchmark, 
ensuring portability of our results to real-world upscaling by using established versions of quality control, preprocessing and evaluation. In contrast to FX, we fix and release the benchmark to enable systematic method comparison, introduce progressively harder extrapolation scenarios, and place explicit emphasis on tail performance.
Recent ML-facing resources include the benchmarks datasets of \citet{fortier2025carbonsensemultimodaldatasetbaseline} and \citet{benchmarkET2025li}; these provide useful datasets, but no fixed data splits or evaluation protocol. 
CarbonBench \citep{rozanov2026carbonbenchglobalbenchmarkupscaling} benchmarks daily fluxes in stratified splits. FLUXtrapolation 
uses hourly data derived from the FX framework, 
making results more directly transferable to current real-world upscaling.
We further consider progressively harder forms of distribution shift and evaluates performance across multiple relevant temporal scales, both of which are not 
considered by CarbonBench.

\section{FLUXNET data and the challenge of upscaling}\label{sec:fluxnet-data}

\paragraph{Data source.} 
FLUXNET is a global network of eddy-covariance sites measuring exchanges of carbon, water, and energy between ecosystems and the atmosphere \citep{pastorello2020fluxnet2015}. In this work, we consider data between 2015 and 2022 from 207 sites 
spanning a wide range of climates, regions, and ecosystem types, though concentrated in North America and Europe. This subset reflects sites available as of 2025 with associated VIIRS-based remote sensing data, which begin in 2015. 

\paragraph{Target variables and covariates.} 
We study three ecosystem fluxes at hourly resolution: net ecosystem exchange of $\mathrm{CO}_2$
(NEE),
which measures net gain or loss of carbon from an ecosystem, gross primary productivity (GPP), which estimates carbon uptake through photosynthesis, and evapotranspiration (ET), which measures water loss through plant transpiration and evaporation from soils and wet surfaces. 
These flux time series are temporally dependent; Figure~\ref{fig:acf-au-cum} (Appendix~\ref{app:data_details}) shows autocorrelation functions for one site.
There are $13$ 
covariates 
for each time point and location,
consisting of meteorological covariates (e.g., 
air temperature (TA),
vapor pressure deficit (VPD)),
site characteristics (e.g., IGBP vegetation type), and satellite-derived variables (e.g., enhanced vegetation index (EVI),
normalized difference water index (NDWI));
we refer to covariates as inputs,
the full list 
is given in Table~\ref{tab:variables}. The covariates follow the FLUXCOM-X framework, including remote-sensing processing \citep{fluxnet-remote-sensing} and quality control \citep{inconsistenciesflux2024jung}; Appendix~\ref{app:data_details} provides details.

\paragraph{Notation and domain definition.} 
We denote the set of FLUXNET sites by $\mathcal{S}$ and the set of years under consideration by $\mathcal{A}:=\{2015,\dots,2022\}$.
The data are observed at hourly resolution, and for each $a\in\mathcal A$, we let $\mathcal T_a$ denote the set of hourly timestamps in year $a$.
For each site $s\in\mathcal{S}$, year $a\in\mathcal{A}$, and time $t\in\mathcal{T}_a$, we observe $(X_t^s,Y_t^s)$, where $X_t^s\in\mathbb R^p$ is the input vector 
($p=13$)
and $Y_t^s\in\mathbb R$ is one target flux variable. 
We treat GPP, NEE, and ET as separate prediction tasks.
We write
$\mathcal D_{s,a}:=\{(X_t^s,Y_t^s): t\in\mathcal T_a\}$
for the data from site $s$ in year $a$, and
$\mathcal D_s:=\bigcup_{a\in\mathcal A}\mathcal D_{s,a}$
for the data from site $s$.
We let $\E$ denote a set of domains.
Depending on the extrapolation scenario (defined in Section~\ref{section:challenge}), 
domains are defined either at the level of sites $s \in \mathcal S$
(that is, train and test domains are defined as subsets of $\mathcal S$)
or site-years
$(s,a) \in \mathcal S \times \mathcal{A}$.

\paragraph{Relevance of the upscaling problem.} 
Much like weather stations monitor meteorological conditions around the world, eddy covariance towers provide estimates of ecosystem function, including GPP, NEE, and ET.
These measurements provide continuous monitoring of ecosystem health and form a foundation for understanding present and future climate and Earth systems. 
Global flux products derived from tower measurements now play an important role in both carbon-cycle science and climate-policy monitoring. 
For example, FLUXCOM-X has been highlighted for European forest-carbon 
monitoring~\citep{migliavaccaSecuringForestCarbon2025a} and used to study Amazon dieback risk~\citep{melnikovaAmazonDieback21st2025} and global shifts in ecosystem water-use efficiency~\citep{greenClimateChangeAltering2025}.

\paragraph{Sources of distribution shift and extrapolation.}
Although FLUXNET is a global eddy-covariance network, tower locations are not chosen by a designed monitoring scheme. 
They reflect local scientific priorities, expertise, resources, and data-sharing practices, producing uneven coverage across geography, climate, and ecosystem type, with dense coverage in Europe and North America but sparse coverage in many target upscaling regions; there is strong covariate shift between observed tower sites and target regions \citep{Stoy2026distrECtowers}. 
Generalization is further complicated by differences in instrumentation, ecosystem representation, data support, processing, and quality control \citep{inconsistenciesflux2024jung}. 
Upscaling can also only use covariates available both at towers and globally, 
but relevant
drivers such as soil and understory properties are partly observed or missing. These hidden variables can change the input--flux relationship, inducing conditional shift. 
We 
expect spatial extrapolation to be harder than temporal extrapolation within sites and for conditional shift to differ across fluxes: while GPP and ET are strongly driven by observed variables such as radiation and vapor pressure deficit, NEE is more influenced by additional, unobserved ecosystem processes.

\section{The FLUXtrapolation benchmark}\label{section:challenge}
\begin{figure}
    \centering
    \includegraphics[scale=1, trim={0 45.5mm 0 0mm}, clip]{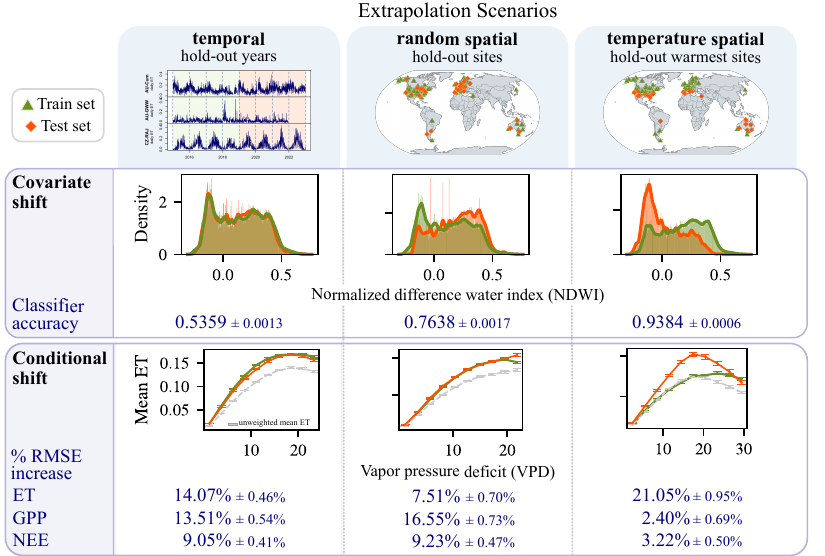}
    \caption{%
    \textit{Overview of the three extrapolation scenarios and the corresponding distribution shifts.} 
    Top row: train--test splits. Middle row: 
    we quantify covariate shift in \(P_X\) by the balanced accuracy of a domain classifier distinguishing training from test inputs in the full covariate space; the plots illustrate this shift through NDWI density plots. 
    Bottom row:
    we quantify 
    conditional shift in $P_{Y\mid X}$ by the percentage increase from held-out training RMSE to test RMSE; the test data are reweighted to match the training marginal over common support (see Section~\ref{sec:quantifying-shifts} for details).
    The plots show the shift in VPD--ET relationship. 
    Covariate shift increases from temporal to spatial to temperature-based extrapolation and conditional shift is present in all scenarios, but does not follow the same ordering.
    }
    \label{fig:benchmark_shifts}
\end{figure}

The FLUXtrapolation benchmark\footnote{
code:
\url{https://github.com/anyafries/FLUXtrapolation}, data: \url{https://doi.org/10.7910/DVN/Q1MPVG}.} on FLUXNET data
poses the task of predicting hourly ecosystem fluxes at held-out sites or site-years from meteorological, satellite-derived, and site-level variables. Each flux (ET, GPP, and NEE) is considered a separate prediction task.
FLUXtrapolation contains
three
extrapolation scenarios with
progressively 
harder to tackle
distribution shift, see Figure~\ref{fig:benchmark_shifts}. 
Each scenario is specified by a choice of
training domains 
(defining the training data) 
and test domains (defining the test data),
both could be a set of sites, for example.
Together, the scenarios range from 
relatively 
mild temporal extrapolation within observed sites to harder spatial and temperature-based extrapolation.
In Section~\ref{sec:quantifying-shifts}, we 
quantitatively 
analyze  
the shifts induced by each scenario 
by estimating the magnitude of the
changes in the marginal and conditional distributions between training and test.
To allow for comparable
model selection, we additionally specify a default validation split for each scenario in Section~\ref{sec:validation}; 
the user, however, may use a different validation set or none at all.

\subsection{Extrapolation scenarios} 

\paragraph{Temporal extrapolation (`temporal').}
Temporal extrapolation  
considers prediction for future years at sites 
that have 
already 
been 
observed during training. 
As
training and test data come from the same sites, 
background knowledge suggests that
this yields a relatively mild extrapolation setting.
In this scenario, we consider only the subset of sites 
$\mathcal{S}^{\mathrm{time}} \subseteq \mathcal  S$ with at least five years of data and observations in 2018, and define domains 
at the level of
site-year pairs. 
Training domains 
consist of 
all 
available 
site-years from 2015–2018 for sites in 
$\mathcal{S}^{\mathrm{time}}$, 
while the test domains 
consist of 
all 
available 
site-years from 2019–2022 for the same sites. 
Formally, 
$\E^\mathrm{time}_\mathrm{train} := \{(s,a) \mid s\in \mathcal{S}^{\mathrm{time}}, 
a \le 2018,\mathcal D_{s,a} \ne \emptyset \}$ 
and 
$\E^\mathrm{time}_\mathrm{test} := \{(s,a) \mid s\in \mathcal{S}^{\mathrm{time}}, 
y\in \mathcal{Y},
a \ge 2019,\mathcal D_{s,a} \ne \emptyset \}$.

\paragraph{%
Random spatial extrapolation (`spatial').}  
Random spatial extrapolation considers prediction at 
sites that are not observed during training.
In this scenario, 
each site is either a training domain or a test domain.
The test sites are randomly selected and, as such, remain broadly similar to the training sites, inducing a moderate shift while retaining substantial overlap in covariate support.
We define disjoint training and test domain sets $\mathcal E_{\mathrm{train}}^{\mathrm{space}}, \mathcal E_{\mathrm{test}}^{\mathrm{space}} \subseteq \mathcal S$, where $\lvert \mathcal E_{\mathrm{test}}^{\mathrm{space}} \rvert = 40$ (roughly 20\% of sites)
and
$\mathcal E_{\mathrm{train}}^{\mathrm{space}} := \mathcal S \setminus \mathcal E_{\mathrm{test}}^{\mathrm{space}}$. 

\paragraph{Temperature spatial extrapolation (`temperature').} Temperature-based spatial extrapolation considers prediction at unseen sites with 
the highest
average annual temperature.
Relative to spatial extrapolation, this compounds generalization to unseen sites with an explicit covariate shift towards warmer conditions. Domains are again defined at the level of sites. Let $\mathcal E_{\mathrm{test}}^{\mathrm{temp}}$ be the set of the 40 sites with highest average annual temperature. The training domains are defined as $\mathcal E_{\mathrm{train}}^{\mathrm{temp}} := \mathcal S \setminus \mathcal E_{\mathrm{test}}^{\mathrm{temp}}$.

\subsection{Analyzing distribution shift for the different extrapolation scenarios}
\label{sec:quantifying-shifts}

To 
analyze 
the differences between
extrapolation scenarios, we quantify the distribution shift in terms of changes in the marginal 
input 
distribution $P_X$ (``covariate shift'') and changes in the conditional relationship $P_{Y\mid X}$ (``conditional shift'', often also called concept shift);
in this work, 
by
covariate shift 
we
refer 
to a shift in $P_X$ and do not imply that $P_{Y\mid X}$ is unchanged. 
For each extrapolation scenario $b$, we pool observations across all training domains and all test domains, letting
$\mathcal D_{\mathrm{train}}^{\,b} := \bigcup_{e \in \mathcal E_{\mathrm{train}}^{\,b}} \mathcal D_e$
and
$\mathcal D_{\mathrm{test}}^{\,b} := \bigcup_{e \in \mathcal E_{\mathrm{test}}^{\,b}} \mathcal D_e$.
We 
now
assess shift between these pooled training and test samples 
visually and quantitatively. 
Figure~\ref{fig:benchmark_shifts} summarizes these comparisons across extrapolation scenarios. 
We measure covariate shift 
by the balanced classifier accuracy distinguishing training from test inputs. This quantity 
increases noticeably from temporal to spatial to temperature-based extrapolation. 
Conditional shift
we quantify 
by the increase in RMSE from training to test after accounting for differences in $P_X$; this diagnostic does not follow the same ordering: for example, for ET, the increase is pronounced for both temporal and temperature-based extrapolation, but smaller for spatial extrapolation. 
The 
scenarios therefore differ 
in the magnitude of distribution shift
and
in whether 
this shift 
arises primarily in $P_X$, in $P_{Y\mid X}$, or in both.
These distinctions make 
our claims about domain generalization more precise: for example, poor performance under mostly $P_{Y\mid X}$ shift (temporal extrapolation) is a different failure mode from poor performance when $P_X$ changes.
The remainder of this subsection defines these two diagnostics and the corresponding visual illustrations 
more precisely.

\paragraph{Marginal shift in $P_X$.}
Figure~\ref{fig:benchmark_shifts}
(``Covariate shift'') shows 
density estimates of
the training and test distributions of the normalized difference water index (NDWI). 
To quantify the multivariate shift, we use the balanced accuracy of a domain classifier 
\citep[see, e.g.,][]{failyloudly2019rabanser}; higher balanced accuracy indicates that training and test inputs are more easily separated and hence indicates a larger shift in $P_X$. 
More precisely,
for each extrapolation scenario $b$, we 
randomly
split 
both
$\mathcal D_{\mathrm{train}}^{\,b}$ and $\mathcal D_{\mathrm{test}}^{\,b}$ into two halves 
and 
join the two first halves (yielding $\D_1^{\,b}$) and the two second halves (yielding $\D_2^{\,b}$). 
We introduce another variable $Z \in\{\mathrm{train},\mathrm{test}\}$ whose value indicates whether a data point belongs to the training or test set. We then
fit a gradient-boosted tree classifier $s:\mathbb{R}^p\to[0,1]$ 
on 
$\D_1^{\,b}$
to distinguish training from test inputs. Here\footnote{%
For readability, we omit the dependence of $s$ on the extrapolation scenario and random split; also, when writing a conditional, we implicitly refer to one version of it.
}, 
$s(x)$ is used as an approximation to
$\mathbb P(Z=\mathrm{test}\mid X=x)$.
We evaluate on the held-out halves 
(i.e., $\D_2^{\,b}$)
and report the balanced accuracy $\pm$~one standard deviation over 10 independent random splits.

\paragraph{Conditional shift in $P_{Y\mid X}$.}
For the quantification of the conditional shift, we use ideas similar to 
the ones discussed by
\citet{cai2025diagnosing}. 
We fit a gradient-boosted tree $f$ on training data, intended to approximate the training conditional mean $x \mapsto \mathbb E_{P^{\mathrm{train}}}[Y \mid X=x]$. 
If 
$P^{\mathrm{train}}_{Y\mid X} = P^{\mathrm{test}}_{Y\mid X}$, then $f$ should incur similar prediction error on training and test once both are evaluated under the same marginal distribution of $X$.
We thus 
use
importance weights $w(x)$ which approximate the ratio of densities $p_X^{\mathrm{train}}(x)/p_X^{\mathrm{test}}(x)$ 
on a region of common support and are 0 elsewhere;
the precise definition of $w$
(and weight clipping)
is described
in the last paragraph of Section~\ref{sec:quantifying-shifts}.
For an extrapolation scenario $b$, we then compare held-out training error with importance-weighted test error,
\begin{equation*}
    \widehat{\mathrm{MSE}}_{\mathrm{test}}^{\,w}(f)
    :=
    \frac{\sum_{(x,y)\in \mathcal D_{\mathrm{test}}^{\,b}} w(x)\,(f(x)-y)^2}
    {\sum_{(x,y)\in \mathcal D_{\mathrm{test}}^{\,b}} w(x)}.
\end{equation*}
When the train and test domains have common support, the density ratio 
estimates
test error under the training marginal distribution of $X$ \citep[e.g.,][]{bickel2009covarshift}.
For each extrapolation scenario $b$, we split $\mathcal D_{\mathrm{train}}^{\,b}$ into three parts for weight estimation, model fitting, and held-out training evaluation, and split $\mathcal D_{\mathrm{test}}^{\,b}$ into two parts for weight estimation and test evaluation.
Figure~\ref{fig:benchmark_shifts} (``Conditional shift'') reports the percentage increase from held-out training RMSE to importance-weighted test RMSE, averaged over 10 random splits; larger increases indicate larger conditional shift. Appendix~\ref{app:rmse-cond-shift} assess the robustness of these metrics.

\paragraph{Visual illustration of conditional shift.}
Figure~\ref{fig:benchmark_shifts} (``Conditional shift'') illustrates conditional shift using the VPD--ET relationship. We restrict VPD to the intersection of its empirical 5th--95th percentile ranges in the training and test data, to ensure common support. We then divide this range into 10 bins and plot the mean ET in the training data with the importance-weighted mean ET in the test data. The weights are computed from the remaining covariates, so that the weighted test curve adjusts for covariate shift outside VPD. A gap between the training curve and the weighted test curve indicates a change in the VPD--ET relationship that is not explained by shifts in the other covariates alone. Appendix~\ref{app:model-based-conditional-shift} confirms the same pattern using a model-based visualization.

\paragraph{Estimating the importance weights.}
For each extrapolation scenario $b$, we estimate the importance weights using the domain classifier $s$ already introduced for quantifying shift in $P_X$. 
We restrict evaluation to train observations with $s(x) > \epsilon$ and test observations with $s(x) < 1-\epsilon$, with $\epsilon=0.1$, 
aiming to retain those observations that have 
non-negligible probability under the other split \citep{crump2009dealing}.
We use the weights
$
w(x):=(1-s(x))/s(x),
$
which estimate the density ratio $p_X^{\mathrm{train}}(x)/p_X^{\mathrm{test}}(x)$ up to a multiplicative constant \citep{bickel2009covarshift, sugiyama2007covariate} for the retained test observations
(formally, for the other observations, we set $w(x):= 0$).
In practice, we 
further
clip these weights at the 99th percentile to stabilize the 
variance of the
importance-weighted estimates 
\citep{ionides2008truncated, orenstein2022winsorized}.
All weighted means and errors are then computed by dividing by the sum of the weights over the corresponding evaluation set.

\subsection{Validation set design}\label{sec:validation} 

Validation data 
may be required for model selection, including hyperparameter tuning, early stopping, and 
model selection.
When the test distribution differs from 
that of
training,
random subsampling of the training data generally does not produce validation data representative of the target setting. In spatiotemporal data, it also ignores spatial and temporal dependence and can 
therefore 
give misleading signals \citep{ploton2020spatial, meyer2021unknownspace, cvStrategy2023sweet}. To be informative, the validation split should resemble the target extrapolation setting while still leaving enough training data to learn a good model and limiting the dependence between training and validation data. As a result, the benchmark effectively compares $($model, validation-strategy$)$ pairs. We do not attempt to evaluate all such strategies here, but emphasize that validation design is an important and reportable part of the evaluation pipeline.
We define one default validation split for each extrapolation scenario $b$ as a reproducible baseline
by partitioning the training data $\D^{\,b}_\mathrm{train}$. For temporal extrapolation, we use 2018 as validation; for spatial and temperature-based extrapolation, we hold out 
20 
randomly sampled
training sites (approximately $10\%$ of all sites).

\subsection{Baselines}\label{sec:baselines}

We evaluate 
several
baselines spanning increasing modeling complexity.
We include linear regression and constant prediction. As 
empirical risk minimization (ERM) approaches, we consider gradient-boosted trees (XGBoost) and multilayer perceptrons (MLPs). 
As an attempt
to account for distribution shift in the input distribution, we include CORAL \citep{baochen2016coral} and an MMD regularized neural network \citep{long2015mmd}. For worst-group robustness 
under distribution shift, 
we include Group DRO \citep{sagawa2019distributionally, hu2018does}. We term the last three methods the domain generalization baselines.
Details on training and hyperparameter tuning are provided in 
Appendix~\ref{app:hyperparams}.

\section{Evaluation protocol}\label{sec:evaluation}

\paragraph{Summarizing the errors over all test domains.}
FLUXNET domains are heterogeneous, so a pooled error can hide substantial variation in model behavior across sites or site-years.
More generally, single aggregate summaries can obscure structured differences in predictive performance across regions or environments \citep[e.g.,][]{meyer2022globalmaps}.
We thus evaluate performance at the level of individual test domains (sites or site-years) rather than only through a single pooled error. 
For each extrapolation scenario $b$ and each test domain $e \in \mathcal E_{\mathrm{test}}^{\,b}$, we compute the RMSE given a prediction function $f$ as
\begin{equation}\label{eqn:rmse}
    \mathrm{RMSE}_e^{\,b}(f)
    :=
    \Big(
    \frac{1}{|\mathcal D_e|}
    \sum_{(x,y)\in \mathcal D_e}
    (f(x)-y)^2
    \Big)^{1/2}.
\end{equation}
We summarize the distribution of these test domain errors with the median and 90th percentile, which, respectively, quantify typical and tail performance.

\paragraph{Multi-scale temporal aggregation.}
Because 
the processes underlying
ecosystem fluxes 
act on different
timescales, evaluation at a single temporal resolution gives only a partial view of model performance. Scientific use cases also often depend on daily, weekly, or seasonal summaries, and strong hourly performance does not necessarily imply accurate aggregated behavior. In line with established practice in ecosystem flux upscaling \citep{tramontana2016, xbase2024}, we evaluate predictions for each test domain at the hourly level and after temporal aggregation to these coarser scales (detailed in Appendix~\ref{app:temporal-agg}). We also consider anomalies (`anom') relative to the mean seasonal cycle and interannual variability (`iav'), which assess 
temporal variation beyond average seasonal patterns. Finally, we evaluate the 
mean annual values across sites (`site-mean'), which assesses whether methods capture persistent differences between sites in flux magnitude.

\paragraph{Summary score.}
To obtain a summary score,
we compute a skill score relative to linear regression. 
Unlike a ranking-based summary, this score is less sensitive to small differences 
between near-tied methods and makes performance more comparable across benchmark cells with different intrinsic difficulty. For each method $m$, let its RMSE for scenario $b$ and temporal scale $c$ be $E_{b,c}^m$ and let $E_{b,c}^0$ be the RMSE for linear regression. We define a method's \emph{skill score} for $(b,c)$ and its \emph{overall skill} as
\begin{equation*}
S^m_{b,c} := 1 - \frac{E_{b,c}^m}{E_{b,c}^0}, 
\qquad 
S^m_{\mathrm{overall}} := \frac{1}{n_{b,c}}\sum_c S_{b,c}^m,
\end{equation*}
where $n_{b,c}$ is the number of $(b, c)$ pairs.
A score of $0$ indicates parity with linear regression, positive values (up to a maximum of one) improvement, and negative values worse performance than linear regression.
As with any baseline-normalized summary, these scores depend on the reference model, so we use them as a compact summary, but not as a stand-alone evaluation metric.

\section{Pilot study 
using
baselines}
We conduct a pilot study of FLUXtrapolation using 
the baselines described in Section~\ref{sec:baselines}.
We show that the extrapolation scenarios together with the chosen evaluation metrics are
able to 
separate 
baselines and 
are
useful for interpreting results of many methods when applied to the benchmark (for example, by revealing differences in tail performance and highlighting failure modes). 
FLUXtrapolation benchmarks
methods' ability to generalize 
across increasing distribution shift, in the tails, and across temporal aggregations.
We present results on ET and comment on the differences across the fluxes later.
Throughout this section, ET values are scaled by 100 for readability.

\subsection{Separability of the baselines}

A simple benchmark set-up could be to perform 
one
temporal 
and one
spatial extrapolation of the fluxes
and 
to evaluate performance with only mean or median 
and only
hourly RMSE across the domains (sites or site-years). Table~\ref{tab:simple-benchmark} (Appendix~\ref{app:add-experimental-results}) would be the resulting benchmark summary; there is little difference between the top four methods.

By introducing 
additional extrapolation scenarios in the benchmark design (Table~\ref{tab:simple-benchmark}, Appendix~\ref{app:add-experimental-results}), the median RMSE increases (from 3.9 to 4.7 and 5.6 for the best 
baseline for temporal, random spatial, and temperature-based 
spatial 
extrapolation, respectively), reflecting increasing difficulty. 
Despite the 
increased
distribution shift, only some additional separation is present between the top
baselines.

\begin{figure}
    \centering
    \includegraphics[scale=1]{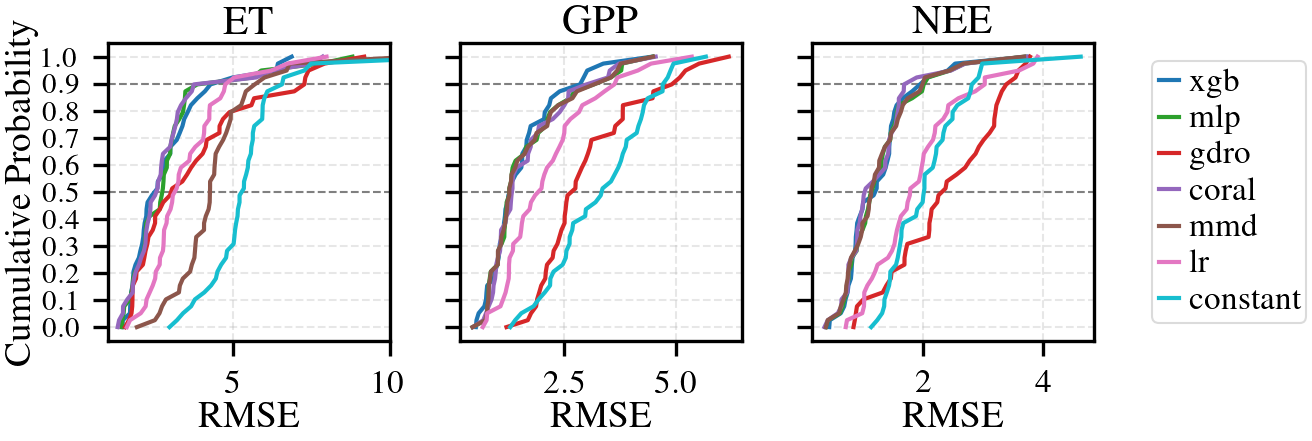} 
    \caption{Cumulative distribution of weekly site-level RMSE for temperature-based extrapolation. Differences between models increase at higher quantiles; the dashed lines mark the 50th and 90th percentiles,
    separation is among the strongest at the 90th percentile. Models separate less for NEE.}
    \label{fig:cdf}
\end{figure}

Considering 90th quantile of RMSE 
(instead of the median) reveals clearer differences between baselines  (Table~\ref{tab:quantile-rmse-et}). 
\begin{table}[t]
    \centering
    \caption{
    \textit{Benchmark summary for the target variable ET based on the 90th percentile of domain-level RMSE.}
    Each cell reports the 90th percentile of domain-level RMSE for a 
    model and scenario--scale pair, see equation~\eqref{eqn:rmse}.
    Cell colors indicate relative performance within each column, with darker green denoting the best value and fading to white at 1.2 times the best value. The final column reports summary score relative to linear regression.
    Model separation increases across the extrapolation scenarios, with CORAL, MLP, and XGBoost consistently 
    among the 
    top
    methods.
    }
    {\small
\setlength{\tabcolsep}{1.4pt}
\begin{tabular}{lcccccc@{\hspace{1.5em}}cccccc@{\hspace{1.5em}}cccccc@{\hspace{1.5em}}c}
\toprule
& \multicolumn{18}{c}{90th percentile of of domain-level RMSE $\downarrow$} & \multirow{3}{*}[-1em]{\rotatebox{90}{Skill score $\uparrow$}} \\
 & \multicolumn{6}{c}{temporal} & \multicolumn{6}{c}{spatial} & \multicolumn{6}{c}{temperature} &  \\
 & \rotatebox{90}{hourly} & \rotatebox{90}{weekly} & \rotatebox{90}{seasonal} & \rotatebox{90}{anom} & \rotatebox{90}{iav} & \rotatebox{90}{site-mean} & \rotatebox{90}{hourly} & \rotatebox{90}{weekly} & \rotatebox{90}{seasonal} & \rotatebox{90}{anom} & \rotatebox{90}{iav} & \rotatebox{90}{site-mean} & \rotatebox{90}{hourly} & \rotatebox{90}{weekly} & \rotatebox{90}{seasonal} & \rotatebox{90}{anom} & \rotatebox{90}{iav} & \rotatebox{90}{site-mean} &  \\
\midrule
coral & \cellcolor[HTML]{63BE7B} 6.4 & \cellcolor[HTML]{74C589} 3.6 & \cellcolor[HTML]{7BC890} 3 & \cellcolor[HTML]{64BE7C} 2.4 & \cellcolor[HTML]{88CD9B} 0.8 & \cellcolor[HTML]{7EC992} 2.2 & \cellcolor[HTML]{7BC88F} 7.6 & \cellcolor[HTML]{9BD5AB} 4 & \cellcolor[HTML]{AEDDBA} 3.6 & \cellcolor[HTML]{63BE7B} 2.6 & \cellcolor[HTML]{63BE7B} 0.84 & \cellcolor[HTML]{63BE7B} 2.7 & \cellcolor[HTML]{99D4A9} 9 & \cellcolor[HTML]{63BE7B} 3.9 & \cellcolor[HTML]{98D4A8} 3.4 & \cellcolor[HTML]{64BE7C} 2.5 & \cellcolor[HTML]{63BE7B} 0.82 & \cellcolor[HTML]{63BE7B} 2.7 & \textbf{0.17} \\
mlp & \cellcolor[HTML]{75C58A} 6.5 & \cellcolor[HTML]{74C589} 3.6 & \cellcolor[HTML]{63BE7B} 2.9 & \cellcolor[HTML]{7AC78E} 2.5 & \cellcolor[HTML]{A1D8B0} 0.82 & \cellcolor[HTML]{94D2A5} 2.3 & \cellcolor[HTML]{63BE7B} 7.3 & \cellcolor[HTML]{8ACE9C} 4 & \cellcolor[HTML]{A9DBB6} 3.5 & \cellcolor[HTML]{6BC182} 2.6 & \cellcolor[HTML]{64BE7C} 0.84 & \cellcolor[HTML]{63BE7B} 2.7 & \cellcolor[HTML]{98D4A8} 9 & \cellcolor[HTML]{74C589} 4 & \cellcolor[HTML]{63BE7B} 3.2 & \cellcolor[HTML]{63BE7B} 2.5 & \cellcolor[HTML]{8ED09F} 0.87 & \cellcolor[HTML]{78C78D} 2.8 & \textbf{0.17} \\
xgb & \cellcolor[HTML]{8FD0A0} 6.7 & \cellcolor[HTML]{63BE7B} 3.5 & \cellcolor[HTML]{65BE7C} 2.9 & \cellcolor[HTML]{64BE7C} 2.4 & \cellcolor[HTML]{CBE9D3} 0.86 & \cellcolor[HTML]{63BE7B} 2.1 & \cellcolor[HTML]{71C487} 7.5 & \cellcolor[HTML]{63BE7B} 3.8 & \cellcolor[HTML]{63BE7B} 3.3 & \cellcolor[HTML]{7CC890} 2.6 & \cellcolor[HTML]{65BE7C} 0.84 & \cellcolor[HTML]{7DC891} 2.8 & \cellcolor[HTML]{63BE7B} 8.4 & \cellcolor[HTML]{B2DFBE} 4.4 & \cellcolor[HTML]{AADBB7} 3.4 & \cellcolor[HTML]{71C487} 2.5 & \cellcolor[HTML]{94D2A4} 0.87 & \cellcolor[HTML]{D1EBD8} 3.1 & \textbf{0.16} \\
gdro & \cellcolor[HTML]{81CA94} 6.6 & \cellcolor[HTML]{69C080} 3.5 & \cellcolor[HTML]{79C78D} 3 & \cellcolor[HTML]{63BE7B} 2.4 & \cellcolor[HTML]{99D4A8} 0.81 & \cellcolor[HTML]{65BE7C} 2.1 & \cellcolor[HTML]{99D4A9} 7.8 & \cellcolor[HTML]{B1DEBD} 4.1 & \cellcolor[HTML]{E7F5EA} 3.8 & \cellcolor[HTML]{7CC890} 2.6 & \cellcolor[HTML]{6FC385} 0.85 & \cellcolor[HTML]{65BE7D} 2.7 & \cellcolor[HTML]{F4FAF6} 10 & \cellcolor[HTML]{FFFFFF} 7.3 & \cellcolor[HTML]{FFFFFF} 7 & \cellcolor[HTML]{97D3A7} 2.6 & \cellcolor[HTML]{A4D9B2} 0.89 & \cellcolor[HTML]{FFFFFF} 5.7 & \textbf{0.02} \\
lr & \cellcolor[HTML]{FFFFFF} 8.1 & \cellcolor[HTML]{FFFFFF} 4.3 & \cellcolor[HTML]{FFFFFF} 4.2 & \cellcolor[HTML]{9BD5AA} 2.6 & \cellcolor[HTML]{63BE7B} 0.76 & \cellcolor[HTML]{FFFFFF} 2.9 & \cellcolor[HTML]{EBF6EE} 8.6 & \cellcolor[HTML]{FFFFFF} 5.5 & \cellcolor[HTML]{FFFFFF} 5.4 & \cellcolor[HTML]{A3D9B1} 2.8 & \cellcolor[HTML]{C4E6CD} 0.94 & \cellcolor[HTML]{FFFFFF} 3.9 & \cellcolor[HTML]{FFFFFF} 10 & \cellcolor[HTML]{FFFFFF} 4.8 & \cellcolor[HTML]{FFFFFF} 4.3 & \cellcolor[HTML]{C9E8D1} 2.8 & \cellcolor[HTML]{9CD5AB} 0.88 & \cellcolor[HTML]{FFFFFF} 3.8 & \textbf{0.00} \\
mmd & \cellcolor[HTML]{FFFFFF} 13 & \cellcolor[HTML]{FFFFFF} 6.6 & \cellcolor[HTML]{FFFFFF} 6.6 & \cellcolor[HTML]{F9FCFA} 2.9 & \cellcolor[HTML]{A3D9B1} 0.82 & \cellcolor[HTML]{FFFFFF} 5.7 & \cellcolor[HTML]{FFFFFF} 12 & \cellcolor[HTML]{FFFFFF} 6.5 & \cellcolor[HTML]{FFFFFF} 5.2 & \cellcolor[HTML]{C2E5CB} 2.9 & \cellcolor[HTML]{64BE7C} 0.84 & \cellcolor[HTML]{FFFFFF} 3.9 & \cellcolor[HTML]{FFFFFF} 14 & \cellcolor[HTML]{FFFFFF} 5.7 & \cellcolor[HTML]{FFFFFF} 5.4 & \cellcolor[HTML]{F9FCFA} 3 & \cellcolor[HTML]{CEEAD5} 0.94 & \cellcolor[HTML]{FFFFFF} 4.3 & \textbf{-0.25} \\
constant & \cellcolor[HTML]{FFFFFF} 13 & \cellcolor[HTML]{FFFFFF} 6.5 & \cellcolor[HTML]{FFFFFF} 6.6 & \cellcolor[HTML]{F9FCFA} 2.9 & \cellcolor[HTML]{A3D9B1} 0.82 & \cellcolor[HTML]{FFFFFF} 5.6 & \cellcolor[HTML]{FFFFFF} 14 & \cellcolor[HTML]{FFFFFF} 6.9 & \cellcolor[HTML]{FFFFFF} 7 & \cellcolor[HTML]{FFFFFF} 3.1 & \cellcolor[HTML]{A9DBB7} 0.92 & \cellcolor[HTML]{FFFFFF} 5.4 & \cellcolor[HTML]{FFFFFF} 16 & \cellcolor[HTML]{FFFFFF} 6.6 & \cellcolor[HTML]{FFFFFF} 6 & \cellcolor[HTML]{FFFFFF} 3.2 & \cellcolor[HTML]{FFFFFF} 1.1 & \cellcolor[HTML]{FFFFFF} 5.4 & \textbf{-0.37} \\
\bottomrule
\end{tabular}
}
    \label{tab:quantile-rmse-et}
\end{table}
For example, the spread in the 90th quantile of hourly RMSE between the best and fourth baseline is $>\!19$\% for temperature-based extrapolation.
This difference is 
amplified by the proposed temporal aggregation in the evaluation protocol, with 
site-mean, weekly, and seasonal scales providing clearer separation.
While we propose using the median and 90th quantile of RMSE,
each extrapolation scenario and temporal aggregation implies a distribution of errors over domains (sites or site-years); Figure~\ref{fig:cdf} shows the cumulative distributions for weekly error for temperature-based extrapolation. 
Methods that are relatively close 
in 
median performance
(e.g., for ET, MLP and Group DRO) may diverge more at higher quantiles.
This highlights 
that median hourly performance does not imply strong performance in the tails or across different temporal aggregations, underscoring evaluation in these cases. 
While separability 
follows from
our evaluation choices, they are chosen foremost to assess robustness 
in a way that is physically meaningful and relevant (see Section~\ref{sec:evaluation}).

\subsection{Preview of
potential
insight
generated by
FLUXtrapolation}

\paragraph{A ready-to-use product.}
From a practical perspective, the benchmark identifies strong baselines.
In 
our
pilot study, XGBoost, MLP, and CORAL consistently rank among the top performers;
XGBoost is also
widely used in current global flux upscaling products.
At the same time, the benchmark provides a basis for evaluating whether more advanced methods can offer consistent 
improvements over these strong baselines and potentially be 
integrated
in
one of the widely used
flux upscaling products.

\paragraph{Revealing relevant failure modes.}
The combination of temporal aggregation and tail-focused evaluation reveals failure modes of the baselines that 
may not be detected
by median hourly evaluation.
For example, consider the 90th quantile of RMSE of inter-annual variability under spatial extrapolation (column `iav' in the second block of Table~\ref{tab:quantile-rmse-et}). 
Here, all baselines perform similarly to the simplest predictor, the constant function.
The baselines additionally approach the constant function or linear regression for the RMSE of IAV and anomalies under temporal extrapolation (though, here, IAV has limited power due to the small number of test years).
Interannual-variability and anomalies
are important for understanding ecosystem dynamics beyond average seasonal behavior and
improving performance here remains a key challenge. Temporal methods are a promising direction \citep[e.g.,][]{kraft2025sequential}.

\paragraph{Challenges in leveraging shift information for improved performance.}
In Section~\ref{sec:quantifying-shifts}, we quantify the type and magnitude of distribution shift in the extrapolation scenarios. 
We now examine whether domain generalization (DG) baselines
, which 
account for distribution shifts, 
lead to improved performance.
For temporal extrapolation, where covariate shift is minimal and conditional shift is moderate, the DG baselines perform similarly to ERM baselines. Even though Group DRO minimizes the worst-case training loss, this does not translate to improved (test) tail RMSE.
For spatial extrapolation, which consists primarily of covariate shift, MMD, designed for covariate shift, drops sharply in performance.
Only CORAL consistently matches the performance of ERM baselines.
Improvements from the baseline DG methods are thus limited and do not align cleanly with the type or magnitude of distribution shift;
this is consistent with findings from prior DG benchmarks, which show that DG methods often fail to reliably outperform ERM baselines under real-world distribution shifts \citep{gulrajani2021domainbed, tableshift2023gardner, koh2021wilds}.
While 
the proposed 
FLUXtrapolation
benchmark
makes shift structure explicit, effectively leveraging this information for model design remains open. 
A related question is whether performance can be gained by a 
well-selected 
validation set. Gaining insight on this could even inform the design of future challenges (as our test set can be viewed as a validation set for the towerless locations).

\subsection{Differences across the fluxes}
The three fluxes, ET, GPP, and NEE, 
measure different ecosystem properties (see Section~\ref{sec:fluxnet-data}). As such, it is unsurprising that the pilot study for ET does not carry over 
to the other fluxes; 
Appendix~\ref{app:add-experimental-results} provides the median and 90th quantile of domain-level RMSE for all three fluxes.
Generally, no baseline dominates across the fluxes and temporal scales.
The results for GPP are qualitatively similar to those of ET. 
However, the 
conclusions differ
for NEE, where the baselines are less separated (for example, see Figure~\ref{fig:cdf}). 
These differences
can be explained by the biological processes that drive the fluxes:
NEE is more strongly influenced by unobserved processes (see Section~\ref{sec:fluxnet-data}).

\section{Discussion, limitations, outlook}\label{sec:discussion}
We introduced FLUXtrapolation, a domain-science motivated benchmark for ecosystem flux prediction under progressively harder extrapolation. Due to the realistic distribution shifts and the relevance of the scientific problem, 
it could support progress both in machine learning for domain generalization and in ecosystem flux upscaling.
It is subject to the following limitations. 
Our shift diagnostics and evaluation summaries are informative but imperfect: distances in observed feature space need not reflect the features most relevant for prediction, conditional-shift comparisons rely on regions of common support, and metrics such as RMSE do not fully account for differences in intrinsic variability across sites, for which regret-style evaluations or more robust error metrics (e.g.\ median absolute error) may be an interesting avenues.

The extrapolation scenarios in FLUXtrapolation are not meant to reproduce the full upscaling problem exactly. Instead, they provide controlled stress tests of shifts that are scientifically plausible in ecosystem flux upscaling, especially toward climates and ecosystem conditions that are poorly represented in the 
FLUXNET network. This is relevant, for example, 
in regions such as the tropics or along wet and dry extremes, where extrapolation is unavoidable. 
Such stress tests are increasingly needed as 
global ecosystems are changing into new regimes due to climate change and human impacts, and because measurement sites will continue to come on and off line 
over time 
\citep{papale2020ideas}.

A next test will be to evaluate 
strong methods
on new sites from the enlarged FLUXNET network and assess whether the stress tests 
were sufficient to identify the more robust methods.
It remains to see
whether external information such as  
foundation models \citep{brown2025alphaearth, feng2026tessera, hollmann2023tabpfn}
can help reduce the effect of hidden variables.
Finally, many other problems in Earth system science are rich in time but sparse in space and require prediction 
where observations are unavailable, including
plant traits \citep{planttraits}, 
tree water use 
\citep{treesap}, 
and soil properties \citep{soilproperties}.
FLUXtrapolation's design and the lessons learned may therefore transfer beyond ecosystem fluxes to a range of environmental problems.

\nocite{ameriflux2025_1881566,ameriflux2025_1881567,ameriflux2026_2006962,ameriflux2026_2469451,ameriflux2025_2469436,ameriflux2025_1818371,ameriflux2025_1881590,ameriflux2026_1881589}
\nocite{ameriflux2026_2204876,ameriflux2026_1881588,ameriflux2026_2229389,grillenburg2024_handle,hetzdorf2024_handle,klingenberg2024_handle,tharandt2024_handle,oberbrenburg2024_handle}
\nocite{ameriflux2026_1881568,ameriflux2025_1854366,ameriflux2026_1871141,ameriflux2025_1832164,ameriflux2025_2469445,ameriflux2026_2007170,ameriflux2026_2469470,ameriflux2026_1881585}
\nocite{ameriflux2026_2316062,chamau2024_handle,grignon2024_handle,toulouse2024_handle,kresinupacova2021_handle,siikaneva2021_handle,maasmechelen2021_handle,sanrossore22022_handle}
\nocite{montebondone2021_handle,lavarone2021_handle,gebesee2021_handle,lamasquere2021_handle,albuera2021_handle,majadasdeltietarsouth2021_handle,bilos2022_handle,lonzee2021_handle}
\nocite{vielsalm2021_handle,castelporziano22021_handle,lanzhot2021_handle,llanodelosjuanes2021_handle,dorinne2021_handle,brasschaat2021_handle,rosinedal32021_handle,borgocioffi2021_handle}
\nocite{hohesholz2021_handle,fontblanche2021_handle,qvidja2021_handle,hyltemossa2021_handle,norunda2021_handle,svartberget2021_handle,soroe2022_handle,hyytiala2021_handle}
\nocite{loobos2019_handle,aguamarga2021_handle}
\nocite{renon2021_handle}
\nocite{yatir2021_handle}
\nocite{clara2022_handle,lanna2020_handle,ameriflux2026_1881594,ameriflux2025_1881565}
\nocite{stordalengrassland2024_handle,rajec2024_handle,stitna2024_handle,ameriflux2024_2204879,ameriflux2024_2229386,ameriflux2026_1881577,ameriflux2026_2006975,trebon2024_handle}
\nocite{ameriflux2026_2204881,ameriflux2025_1881570,ameriflux2025_1881571,ameriflux2025_1881572,ameriflux2026_1881569,ameriflux2026_1871138,ameriflux2026_1881583,ameriflux2026_2007175}
\nocite{mejusseaume2024_handle,ameriflux2026_1881587,ameriflux2026_1881591,ameriflux2026_1881592,ameriflux2026_2229387,ameriflux2026_2229377,ameriflux2026_2229378,ameriflux2026_1871136}
\nocite{skjern2024_handle,boscofontana2024_handle,ameriflux2026_1881582,ameriflux2026_2469472,ameriflux2026_2469473,ameriflux2023_2204882,ameriflux2026_1881597,bibaibog2024_handle}
\nocite{ameriflux2026_1881573,ameriflux2026_1881574,ameriflux2026_1881575,puechabon2024_handle,reinshof2024_handle,hainich2024_handle,ameriflux2025_1881564,varrio2024_handle}
\nocite{lettosuo2024_handle,padul2024_handle,ameriflux2026_1854371,ameriflux2021_1818370,ameriflux2024_1854368,ameriflux2026_2204871,ameriflux2026_2469442,estreesmonsa282024_handle}
\nocite{ameriflux2026_1832161,ameriflux2026_1984572,ameriflux2026_1984573,ameriflux2026_2229391,ameriflux2026_2229392,ameriflux2026_1902832,ameriflux2026_2204880,ameriflux2016_1440200}
\nocite{ameriflux2026_2469476,auchencorthmoss2024_handle,ameriflux2025_1881598,ameriflux2025_1985439,ameriflux2025_1985442,ameriflux2025_1985443,ameriflux2025_1985448,ameriflux2025_1985455}
\nocite{ameriflux2025_2229407,ameriflux2026_1985436,ameriflux2026_1985438,ameriflux2026_1985445,ameriflux2026_1985452,ameriflux2026_1985456,ameriflux2026_1985457,ameriflux2026_1985458}
\nocite{ameriflux2026_2229408,ameriflux2026_2229411,ameriflux2026_2229412,muntatschinigmeadow2024_handle,muntatschinigpasture2024_handle,ameriflux2026_2204872,ameriflux2026_1902837,ameriflux2026_2469450}
\nocite{ameriflux2026_1854369,ameriflux2026_2006970,ameriflux2026_2469446,ameriflux2026_2006964,ameriflux2024_2469453,ameriflux2026_1881581,alicespringsmulga2014_handle,ameriflux2016_1440196}
\nocite{ameriflux2025_2469441,ameriflux2026_1832158,lison2024_handle,ameriflux2024_2469434,ameriflux2026_1871135,ameriflux2025_1871134,ameriflux2026_2006969,ameriflux2026_1832160}
\nocite{selhausenjuelich2024_handle,rollesbroich2024_handle,ameriflux2026_2229383,ameriflux2026_2229384,ameriflux2026_2204874,ameriflux2026_2204873,ameriflux2026_2204875,ameriflux2026_2229376}
\nocite{ameriflux2026_1984574,ameriflux2025_1984575,ameriflux2026_2204877,ameriflux2026_2475756,ameriflux2026_1871144,ameriflux2025_1832163,ameriflux2025_2469439,ameriflux2025_2469440}
\nocite{arcadinoeleprigionette2024_handle,ameriflux2022_1854365,ameriflux2026_1871140,aurade2024_handle,ameriflux2020_2g60zhak,ameriflux2025_1902821,ameriflux2026_1902822,ameriflux2026_1902838}
\nocite{ameriflux2026_2229388,ameriflux2026_1832165,ameriflux2026_2229390,fyodorovskoye2024_handle,fyodorovskoyedrysprucestand2024_handle,ameriflux2024_1881584,ameriflux2026_2204878,ameriflux2026_1881563}
\nocite{ameriflux2025_1832154,ameriflux2026_1854370,torgnonld2024_handle,voulundgaard2024_handle,bilykrizforest2024_handle}
\begingroup
\small
\raggedright
\bibliographystyle{apalike}
\bibliography{bib}
\endgroup

\clearpage
\appendix

\startcontents[appendices]
\section*{Contents of the Appendix}
\printcontents[appendices] 
  {l}                      
  {1}                      
  {\setcounter{tocdepth}{1}} 
\vspace{10mm}

\section{Further data details}
\label{app:data_details}

\subsection{FLUXNET preprocessing and quality control}
\label{app:fluxnet_preprocessing}

\begin{figure}[htbp]
    \centering

    \includegraphics[width=1.0\textwidth, trim={0cm 0mm 0cm 0cm}, clip]{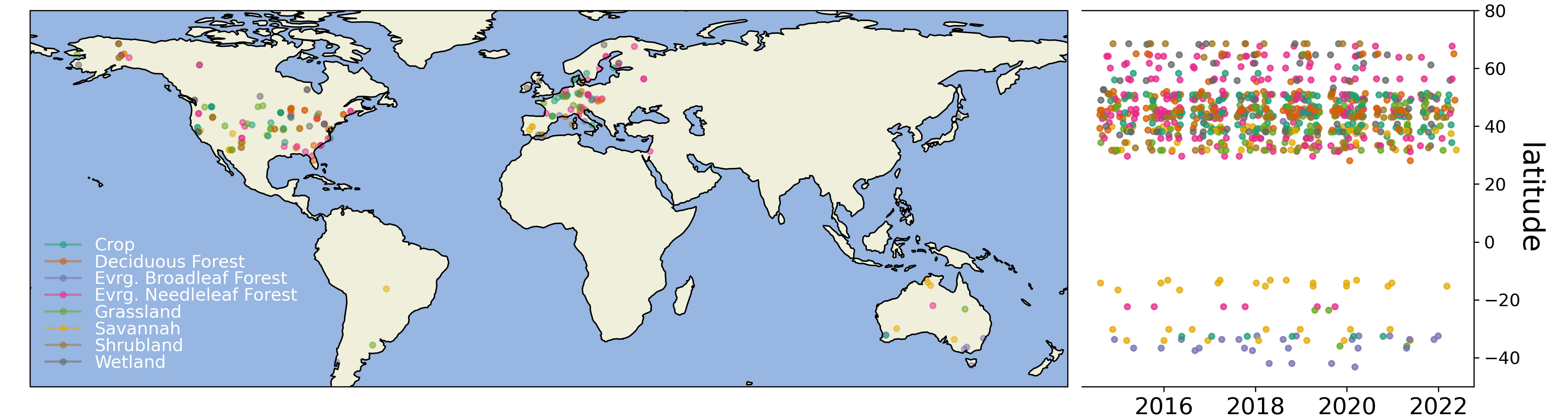}

    \caption{Overview of tower locations in space and time. Points colored corresponding to aggregated PFT classes for clarity.}
    \label{fig:MapTime}
\end{figure}

All data processing and quality control procedures follow established protocols from the FLUXCOM-X pipeline, enabling seamless integration with global flux modeling and benchmarking efforts. 
The underlying FLUXNET data are publicly available via the Dataverse repository 
(\url{https://doi.org/10.7910/DVN/Q1MPVG})
under the CC-BY 4.0 license.
Figure~\ref{fig:MapTime} gives an overview of the spatial and temporal distribution of sites included and Table~\ref{tab:variables} lists the input variables, metadata fields, and target variables used in the benchmark, together with their units and brief descriptions.

\paragraph{Eddy Covariance Data}
Hourly eddy covariance flux measurements were sourced from 207 global sites with a temporal span between 2015 and 2022. FLUXNET data are collected, processed, and quality-controlled by individual site teams, then harmonized using the ONEFlux data processing pipeline (Pastorello et al., 2020), including standardized gap-filling via marginal distribution sampling (MDS), consistent quality flags, and carbon flux partitioning. As distributed FLUXNET data can have either a half-hourly or hourly temporal resolution, the original data was aggregated to a common hourly time step. Target variables include net ecosystem exchange (NEE), gross primary productivity (GPP), and evapotranspiration (ET). GPP was derived using nighttime partitioning method (Reichstein et al., 2005), while ET was computed from latent heat flux.

\paragraph{Meteorological Data}
Site-level meteorological variables, air temperature (TA), vapor pressure deficit (VPD), and incoming shortwave radiation, are also distributed as gap-filled using MDS. Longer gaps in meteorological data which result in poor gap-filling from MDS were filled with a bias corrected estimate from the nearest pixel of the global ERA5 reanalysis product. Potential incoming shortwave radiation (top-of-atmosphere maximum) was computed for each hour.

\paragraph{Satellite Earth Observation:}
For each site, we construct VIIRS-based predictors rather than using the published X-BASE 
products, which are based on MODIS remote sensing. Specifically,
$1\,\mathrm{km}^{2}$ radius cutouts around each tower were processed using a dynamic quality control procedure (Walther \& Besnard et al., 2022) that removes cloud, snow, and water pixels and enforces physical bounds on indices. All remote sensing is also gap filled to prevent a clear sky bias in the training data. Enhanced vegetation index (EVI), near-infrared reflectance (NIRv), and normalized difference water index (NDWI, based on the SWIR2 band), were computed from reflectance values from the combined VNP43MA and VJ143MA products. Land surface temperature (LST) was derived from two daily overpasses (day and night), treated as independent predictors, derived from the VJ121A1 and VNP21A1 products.

\paragraph{Quality Control}
A multi-stage quality control protocol ensures data integrity both within and across sites. The quality information is encoded into a single binary quality flag which indicates all targets and features are of sufficient quality for training. At the hourly level, only time points with at least one half-hour measurement (original FLUXNET quality flag of 0) were retained. At the daily level, consistency checks were applied across meteorological and flux variables, including cross-variable relationships and inter-site coherence, according to Jung et al 2020. Days failing these tests were marked as non-training data quality. The final dataset comprises high-quality, temporally aligned, and physically consistent inputs and targets, enabling robust evaluation of model generalization and extrapolation performance across diverse ecosystems and temporal scales.

\begin{table}[ht]
\centering
\small
\setlength{\tabcolsep}{4pt}
\begin{tabularx}{\linewidth}{lllX}
\toprule
\hspace{1mm} & \textbf{Variable} & \textbf{Units} & \textbf{Description} \\
\midrule
\multicolumn{4}{l}{\textit{Time and metadata}} \\
& time & -- & Timestamp of the hourly observation. \\
& year & -- & Calendar year. \\
& tower\_lon & degrees & Tower longitude. \\
& tower\_lat & degrees & Tower latitude. \\
& qc\_mask & -- & Quality-control indicator used to filter observations. \\
\multicolumn{4}{l}{\textit{Meteorological variables}} \\
& TA & $^\circ$C & Air temperature. \\
& VPD & hPa & Vapour pressure deficit. \\
& SW\_IN & W\,m$^{-2}$ & Incoming shortwave (SW) radiation. \\
& SW\_IN\_POT & W\,m$^{-2}$ & Potential incoming SW radiation. \\
& SW\_IN\_POT\_daily & W\,m$^{-2}$ & Daily mean potential incoming SW radiation. \\
& dSW\_IN\_POT & W\,m$^{-2}$\,hr$^{-1}$ & Change in potential incoming SW radiation. \\
& dSW\_IN\_POT\_daily & W\,m$^{-2}$\,d$^{-1}$ & Daily change in potential incoming SW radiation. \\
\multicolumn{4}{l}{\textit{Satellite-derived variables}} \\
& LST\_Day & K & Daytime land surface temperature. \\
& LST\_Night & K & Night-time land surface temperature. \\
& EVI & -- & Enhanced vegetation index. \\
& NIRv & -- & Near-infrared reflectance of vegetation. \\
& NDWI\_SWIR2 & -- & Normalized difference water index based on SWIR2. \\
\multicolumn{4}{l}{\textit{Plant functional type}} \\
& PFT & -- & Plant functional type, one-hot encoded using the classes CRO, CSH, CVM, DBF, DNF, EBF, ENF, GRA, MF, OSH, SAV, WAT, WET, and WSA. \\
\multicolumn{4}{l}{\textit{Targets}} \\
& NEE & $\mu mol\,CO_{2}$\,m$^{-2}$\,s$^{-1}$ & Net ecosystem exchange. \\
& GPP & $\mu mol\,CO_{2}$\,m$^{-2}$\,s$^{-1}$ & Gross primary productivity. \\
& ET & mm\,hr$^{-1}$ & Evapotranspiration. \\
\bottomrule
\end{tabularx}
\caption{Variables used in the dataset. Inputs are grouped by category. Plant functional type (PFT) is represented through one-hot encoding.}
\label{tab:variables}
\end{table}

\subsection{Autocorrelation structure of the flux time series}
\label{app:acf}

\begin{figure}[ht]
    \centering
    \includegraphics[scale=1]{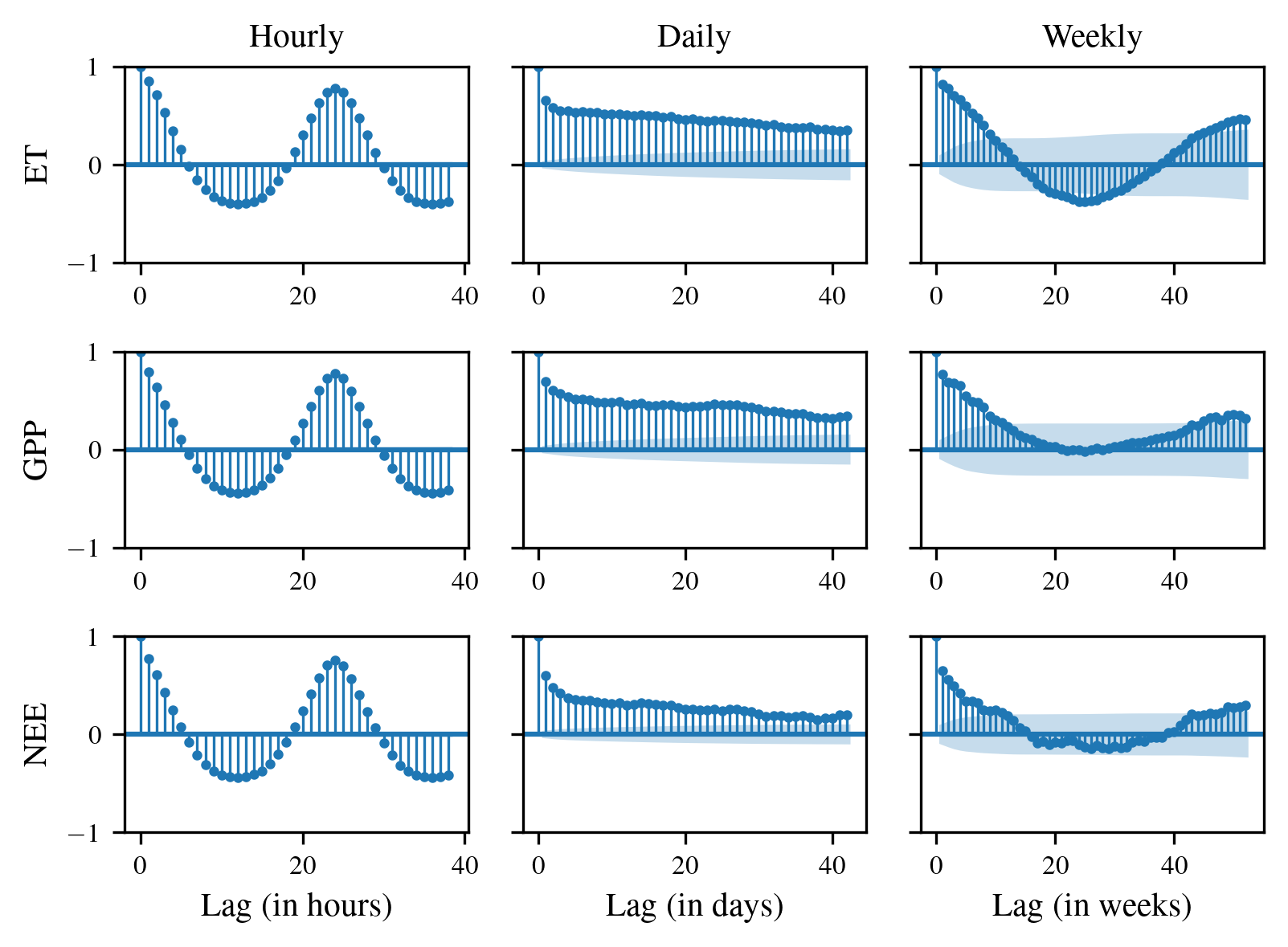}
    \caption{Autocorrelation functions for ET, GPP, and NEE at site AU-Cum, shown at hourly, daily, and weekly resolution. The plots illustrate substantial temporal dependence in the flux time series across multiple timescales.}    
    \label{fig:acf-au-cum}
\end{figure}

To illustrate the temporal dependence present in the flux time series, Figure~\ref{fig:acf-au-cum} shows autocorrelation functions for one site, AU-Cum, for all three target fluxes. We report autocorrelation at hourly, daily, and weekly resolution to illustrate the benchmark data are temporally structured rather than approximately independent over time, which motivates both the temporal extrapolation setting and evaluation across multiple temporal scales.

\section{Robustness checks for conditional-shift diagnostics}

\subsection{Alternative RMSE-based conditional-shift diagnostics}\label{app:rmse-cond-shift}

The main-text diagnostic evaluates both training and test RMSE under the training marginal distribution of $X$. Reversing the reference marginal and evaluating both errors under the test distribution would require weights proportional to $p_X^{\mathrm{test}}(x)/p_X^{\mathrm{train}}(x)$, which were large and unstable in our experiments (not shown). As a complementary check, we therefore instead evaluate both errors under a shared-support reference distribution $S_X$, following \citet{cai2025diagnosing}.

The shared distribution places mass on regions where both training and test covariate distributions have support.
Let $\alpha$ denote the proportion of observations in the pooled sample that come from the test distribution.
We consider the following density as the shared density
\begin{equation*}
p_X^\mathrm{shared}(x)
\propto
\frac{p_X^{\mathrm{train}}(x)\,p_X^{\mathrm{test}}(x)}
{(1-\alpha)p_X^{\mathrm{train}}(x)+\alpha p_X^{\mathrm{test}}(x)} .
\end{equation*}
Let $s(x) \approx \mathbb P(Z=\mathrm{test}\mid X=x)$ be the output of the domain classifier. We weight training and test observations by, 
respectively,
\begin{equation*}
w_{\mathrm{train}\to S}(x) := \frac{s(x)}{\alpha},
\qquad
w_{\mathrm{test}\to S}(x) := \frac{1-s(x)}{1-\alpha}.
\end{equation*}
For the same fitted model $f_{\mathrm{train}}$, we then compare the weighted training and test RMSEs under this shared marginal,
\begin{equation*}
\widehat{\mathrm{RMSE}}_{\mathrm{train}}^{\,S}(f_{\mathrm{train}})
:=
\Bigg(
\frac{\sum_{(x,y)\in \mathcal D_{\mathrm{train}}} 
w_{\mathrm{train}\to S}(x)\,(f_{\mathrm{train}}(x)-y)^2}
{\sum_{(x,y)\in \mathcal D_{\mathrm{train}}} w_{\mathrm{train}\to S}(x)}
\Bigg)^{1/2},
\end{equation*}
and analogously for $\widehat{\mathrm{RMSE}}_{\mathrm{test}}^{\,S}(f_{\mathrm{train}})$. 
We 
now consider
the percentage increase from $\widehat{\mathrm{RMSE}}_{\mathrm{train}}^{\,S}$ to $\widehat{\mathrm{RMSE}}_{\mathrm{test}}^{\,S}$.

\paragraph{Sample splitting.}
To preserve independence between weight estimation, model fitting, and evaluation, we use disjoint splits throughout. 
The training data are split into three parts for weight estimation, model fitting, and held-out evaluation, while the test data are split into two parts for weight estimation and evaluation. 
All reported results are averaged over 10 independent random splits and shown with one standard deviation.

\paragraph{Summary.}
Table~\ref{tab:conditional_shift_shared_marginal} shows that the shared-marginal diagnostic gives broadly similar conclusions to the main training-marginal diagnostic. The absolute magnitudes change, as expected because the two diagnostics evaluate risk under different reference marginals, but the qualitative patterns are largely preserved. For GPP and NEE, the shared-marginal diagnostic assigns relatively more conditional shift to temperature extrapolation than the training-marginal diagnostic, suggesting that part of the train-marginal comparison is sensitive to the chosen reference distribution. 
\begin{table}[t]
\centering
\small
\setlength{\tabcolsep}{6pt}
\begin{tabular}{lccc}
\toprule
& \multicolumn{3}{c}{\textbf{Extrapolation scenario}} \\
\cmidrule(lr){2-4}
& \textbf{temporal} & \textbf{spatial} & \textbf{temperature} \\
\midrule
\multicolumn{4}{l}{\textbf{\% RMSE increase (weighted to }$P_X^{\mathrm{train}}$\textbf{)}} \\
\addlinespace[1pt]
ET  & 14.07\% \tiny$\pm$ 0.46\% & 7.51\% \tiny$\pm$ 0.70\% & 21.05\% \tiny$\pm$ 0.95\% \\
GPP & 13.51\% \tiny$\pm$ 0.54\% & 16.55\% \tiny$\pm$ 0.73\% & 2.40\% \tiny$\pm$ 0.69\% \\
NEE & 9.05\% \tiny$\pm$ 0.41\% & 9.23\% \tiny$\pm$ 0.47\% & 3.22\% \tiny$\pm$ 0.50\% \\
\addlinespace[3pt]
\midrule
\multicolumn{4}{l}{\textbf{\% RMSE increase (weighted to shared-support }$S_X$\textbf{)}} \\
\addlinespace[1pt]
ET  & 13.48\% \tiny$\pm$ 0.48\% & 5.82\% \tiny$\pm$ 0.62\% & 17.32\% \tiny$\pm$ 0.87\% \\
GPP & 12.02\% \tiny$\pm$ 0.50\% & 11.23\% \tiny$\pm$ 0.55\% & 8.00\% \tiny$\pm$ 0.80\% \\
NEE & 7.68\% \tiny$\pm$ 0.38\% & 4.67\% \tiny$\pm$ 0.44\% & 6.99\% \tiny$\pm$ 0.69\% \\
\bottomrule
\end{tabular}
\caption{Conditional-shift diagnostics across extrapolation scenarios for models fitted on the training domains. We report the percentage increase in RMSE under two reference marginals: the training marginal $P_X^{\mathrm{train}}$ and the shared-support marginal $P_X^\mathrm{shared}$. Values are mean $\pm$ standard deviation over ten repeated random splitting plus refitting.}
\label{tab:conditional_shift_shared_marginal}
\end{table}

\subsection{Model-based visualization of conditional shift}\label{app:model-based-conditional-shift}

\begin{figure}[htbp]
    \centering
    \begin{subfigure}[b]{0.25\textwidth}
        \hspace*{-8mm}
        \includegraphics[scale=1, trim={0cm 3mm 0cm 0cm}, clip]{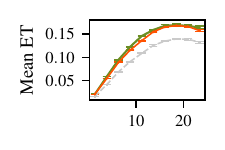}

        {\hspace*{14mm}\small VPD}

        \caption{}
        \label{fig:first}
    \end{subfigure}
    \begin{subfigure}[b]{0.25\textwidth}
        \hspace*{-8mm}
        \includegraphics[scale=1, trim={0cm 3mm 0cm 0cm}, clip]{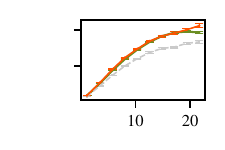}

        {\hspace*{14mm}\small VPD}
        
        \caption{}
        \label{fig:second}
    \end{subfigure}
    \begin{subfigure}[b]{0.25\textwidth}
        \hspace*{-8mm}
        \includegraphics[scale=1, trim={0cm 3mm 0cm 0cm}, clip]{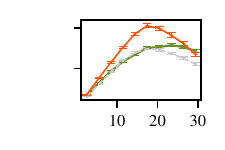} \\
        {\hspace*{14mm}\small VPD}
        
        \caption{}
        \label{fig:third}
    \end{subfigure}
    \begin{subfigure}[b]{0.15\textwidth}
        \raisebox{0pt}[\height][\height]{%
          \begin{minipage}[c][4cm][c]{\textwidth}
            \centering
            \includegraphics[width=\textwidth]{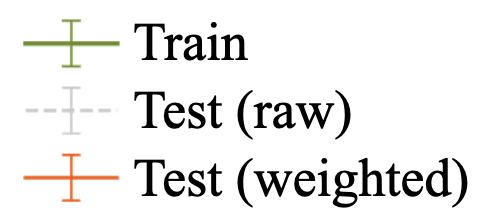}
          \end{minipage}%
        }
        \label{fig:legend}
    \end{subfigure}
    \caption{Model-based illustration of conditional shift in the VPD--ET relationship across extrapolation scenarios. For each scenario, we fit separate gradient-boosted regression trees on the training and test domains, bin VPD on the common-support range, and plot the mean fitted response in each bin. The test curve is importance-weighted to adjust for covariate shift outside VPD. A gap between the training curve and the weighted test curve indicates a change in the VPD--ET relationship that is not explained by shifts in the other covariates alone.}
    \label{fig:model-based-conditional-shift}
\end{figure}

Figure~\ref{fig:model-based-conditional-shift}
complements the illustrative conditional-shift plot of the paragraph ``Visual illustration of conditional shift'' in Section~\ref{sec:quantifying-shifts}.
Instead of plotting binned means of observed ET, we 
now
plot binned means of model-based response curves. For each extrapolation scenario, we split the training and test domains into three parts for, 
respectively,
regressor fitting, evaluation, and domain-classifier fitting (yielding $\mathcal D_1^t, \mathcal D_2^t, \mathcal D_3^t$ for $t\in\{\mathrm{train}, \mathrm{test}\}$).
On each of $\mathcal D_1^\mathrm{train}$ and $\mathcal D_1^\mathrm{test}$, we fit a gradient-boosted regression tree, denoted by $f_{\mathrm{train}}$ and $f_{\mathrm{test}}$; each is intended to approximate the conditional mean of ET given the covariates in its own domain.
We predict ET using $f_{\mathrm{train}}$ and $f_{\mathrm{test}}$ on the respective evaluation splits, $\mathcal D_2^\mathrm{train}$ and $\mathcal D_2^\mathrm{test}$.
To visualize the conditional relationship, we restrict VPD to the intersection of its empirical 5th--95th percentile ranges in $\mathcal D_2^\mathrm{train}$ and $\mathcal D_2^\mathrm{test}$, and divide this range into 10 bins, as in the paragraph ``Visual illustration of conditional shift'' in Section~\ref{sec:quantifying-shifts}.
Within each bin, the training curve is the mean of $f_{\mathrm{train}}(x)$ over the training points falling in that bin, and the test curve is the importance-weighted mean of $f_{\mathrm{test}}(x)$ over the test points in that bin, with weights normalized to sum to one within the bin.
The importance weights are calculated from a classifier $s$ fit to distinguish $\mathcal D_3^\mathrm{train}$ and $\mathcal D_3^\mathrm{test}$, as in the paragraph ``Estimating the importance weights'' in Section~\ref{sec:quantifying-shifts} (details on estimation and weight clipping carry over); here, however, $s$ is fit on all covariates except VPD, so the weights depend only on the remaining covariates and the weighted test curve adjusts for covariate shift outside VPD. A gap between the training curve and the weighted test curve indicates a change in the VPD--ET relationship not explained by shifts in the other covariates alone.
Figure~\ref{fig:model-based-conditional-shift} shows the resulting VPD--ET curves; they 
are similar to
those in Figure~\ref{fig:benchmark_shifts} (``Conditional shift'') of the main text 
and the same qualitative discussion applies.


\section{Temporal aggregation of evaluation metrics}\label{app:temporal-agg}

We evaluate predictions not only at the native hourly resolution but also under several coarser and derived temporal summaries that emphasize different aspects of model performance. 
For each aggregation level, we first aggregate the true and predicted target values and then compute RMSE between these aggregated target--prediction pairs.

To avoid unstable aggregates based on too little data, we impose minimum-contribution thresholds. When aggregating from hourly to daily mean, we require at least 12 valid (quality-controlled) hourly observations.
Coarser aggregates are then computed from these daily values. 
A weekly average is retained only if at least 4 valid daily values are available within that week. 
If either threshold is not met, the corresponding aggregated value is treated as missing and excluded from evaluation. 
For temporal extrapolation, these aggregations are performed at the site--year level, while for spatial and temperature-based extrapolation, they are performed at the site level.

Seasonal aggregation is implemented for each site, which summarizes the typical within-year pattern at each site. 
After converting to daily values, each observation is assigned its day of year, and for each site and day of year we average across all available years. 
This is also known as the the mean seasonal cycle (MSC).
We require at least two valid years per day of year to retain the estimate. 
Leap years are handled by treating day 366 separately.

Anomalies are defined relative to the site-specific MSC: for each daily observation, we subtract the MSC value corresponding to the same site and day of year, separately for targets and predictions. This removes the seasonal pattern and focuses evaluation on deviations from the expected seasonal baseline, i.e., it is a detrended daily value.

To further assess year-to-year fluctuations beyond a site's average level, we compute inter-annual variability (IAV) from yearly aggregates. 
For each site $s$ and year $a$, we compute the yearly mean $\bar{y}_{s,a}$ (retained only if at least $183$ valid days are available) and the site's multi-year mean $\bar{y}_{s}$, yielding one deviation $\mathrm{IAV}_{s,a} := \bar{y}_{s,a} - \bar{y}_{s}$ per site-year. The site mean $\bar{y}_{s}$ is computed from test years only for temporal extrapolation and from all available years for spatial and temperature-based extrapolation. The resulting series captures deviations of each year from that site's typical annual level. RMSE is then computed between the true and predicted IAV series at the site level. For temporal extrapolation, the series has
at most four points per site, and for spatial and temperature-based extrapolation, at most eight.


\section{Experimental details}\label{app:add-experimental-details}

\subsection{Training and hyperparameter tuning}\label{app:hyperparams}

All baselines (see Section~\ref{sec:baselines}) 
are trained using mean squared error and 
hyperparameters are tuned via random search over predefined ranges, with 10 sampled configurations per baseline. 
The baselines are evaluated using the validation protocol described in Section~\ref{sec:validation}, and the best hyperparameters are selected based on mean validation performance.
Further, all models use fixed random seeds for reproducibility.

Linear regression is fit without regularization. 
XGBoost 
hyperparameters include
the number of estimators ($100$--$1000$), tree depth ($3$--$10$), learning rate (log-uniform), and feature subsampling ratios ($0.6$--$1$). 
Early stopping is used based on the validation loss.
The MLP hyperparameters are the number of fully connected layers ($2$--$3$, with ReLU activations), the dropout rate (uniform in $[0,0.5]$), the hidden dimensions (selected from $\{[128,64], [256,128], [512,256,128]\}$),
the learning rate (log-uniform in $[10^{-5}, 10^{-2}]$), and the batch size (selected from $\{512, 1024, 2048\}$). 
All neural models are trained using Adam, with early stopping based on validation loss. 
When mini-batches are constructed, we use stratified-sampling according to sites. 

Domain generalization methods (Group DRO, CORAL, MMD) share the same architecture as the selected MLP and inherit its tuned hyperparameters. 
Additional method-specific parameters are tuned via random search over $5$ configurations:
\begin{itemize}
    \item Group DRO: group weight step size (log-uniform in $[10^{-4}, 10^{-1}]$).
    \item CORAL: regularization strength $\lambda_{\mathrm{coral}}$ (log-uniform in $[10^{-2}, 10^{1}]$).
    \item MMD: regularization strength (log-uniform in $[10^{-2}, 10^{1}]$) with multi-scale Gaussian kernel at fixed bandwidths $\{10^{-3}, 10^{-2}, \ldots, 10^{3}\}$ following the implementation of \citet{tableshift2023gardner} (as also used by \citet{gulrajani2021domainbed}); 
    note, however, that
    the original method \citep{long2015mmd} instead uses multi-kernel MMD with bandwidths centered on the median pairwise distance.
\end{itemize}

Deep models (MLP and domain generalization methods) were trained using a single NVIDIA RTX 4090 GPU (24\,GB), with training time of approximately 1--2 hours per model. Non-deep models (linear regression and XGBoost) were trained on CPUs and completed within under an hour in total (run in parallel). Experiments were conducted on machines with AMD EPYC CPUs and 512\,GB--1\,TB RAM. Total compute for reported experiments was on the order of tens of GPU-hours.

\section{Additional experimental results}\label{app:add-experimental-results}

\subsection{Results under simple evaluation}

Table~\ref{tab:simple-benchmark} reports baseline performance under a simple evaluation based on median hourly RMSE, illustrating that differences between strong baselines are small in this setting.

\begin{table}[ht]
    \centering
    \caption{Median hourly RMSE for ET under temporal and spatial extrapolation. Differences between the top-performing baselines are small in this setting.}
    {\setlength{\tabcolsep}{3pt}
    \begin{tabular}{lccccccc}
    \toprule
    & xgb & mlp & coral & gdro & lr & mmd & constant \\
    \midrule
    temporal & \cellcolor[HTML]{65BF7D} 4.0 & \cellcolor[HTML]{63BE7B} \textbf{3.9} & \cellcolor[HTML]{64BE7C} 4.0 & \cellcolor[HTML]{63BE7B} \textbf{3.9} & \cellcolor[HTML]{FFFFFF} 6.0 & \cellcolor[HTML]{FFFFFF} 9.3 & \cellcolor[HTML]{FFFFFF} 9.3 \\
    spatial  & \cellcolor[HTML]{63BE7B} \textbf{4.7} & \cellcolor[HTML]{82CB95} 4.9 & \cellcolor[HTML]{90D0A1} 5.0 & \cellcolor[HTML]{63BE7B} \textbf{4.7} & \cellcolor[HTML]{FFFFFF} 6.1 & \cellcolor[HTML]{FFFFFF} 6.6 & \cellcolor[HTML]{FFFFFF} 9.5 \\
    \bottomrule
    \end{tabular}
    }
    \label{tab:simple-benchmark}
\end{table}

\subsection{Full results across fluxes}

Tables~\ref{tab:supp-et-median}--\ref{tab:nee-sup-90} report the median and 90th percentile of domain-level RMSE for all three fluxes (ET, GPP, and NEE) across extrapolation scenarios and temporal scales, complementing the ET-focused analysis in the main text.
Across all three fluxes, no single method dominates and the ranking differs by flux and scale. NEE shows the smallest separation between methods, consistent with the role of unobserved drivers discussed in Section~\ref{sec:fluxnet-data}.
\begin{table}[htbp]
    \centering
    \caption{Median of domain-level RMSE for ET $\downarrow$}
    {\small
\setlength{\tabcolsep}{1.4pt}
\begin{tabular}{lcccccc@{\hspace{1.5em}}cccccc@{\hspace{1.5em}}cccccc@{\hspace{1.5em}}c}
\toprule
 & \multicolumn{6}{c}{temporal} & \multicolumn{6}{c}{spatial} & \multicolumn{6}{c}{temperature} & \multirow{2}{*}{\rotatebox{90}{Skill score $\uparrow$}} \\
 & \rotatebox{90}{hourly} & \rotatebox{90}{weekly} & \rotatebox{90}{seasonal} & \rotatebox{90}{anom} & \rotatebox{90}{iav} & \rotatebox{90}{site-mean} & \rotatebox{90}{hourly} & \rotatebox{90}{weekly} & \rotatebox{90}{seasonal} & \rotatebox{90}{anom} & \rotatebox{90}{iav} & \rotatebox{90}{site-mean} & \rotatebox{90}{hourly} & \rotatebox{90}{weekly} & \rotatebox{90}{seasonal} & \rotatebox{90}{anom} & \rotatebox{90}{iav} & \rotatebox{90}{site-mean} &  \\
\midrule
mlp & \cellcolor[HTML]{63BE7B} 3.9 & \cellcolor[HTML]{63BE7B} 1.8 & \cellcolor[HTML]{76C58B} 1.6 & \cellcolor[HTML]{67BF7E} 1.6 & \cellcolor[HTML]{63BE7B} 0.24 & \cellcolor[HTML]{B5E0C0} 0.74 & \cellcolor[HTML]{82CB95} 4.9 & \cellcolor[HTML]{C2E5CC} 2.6 & \cellcolor[HTML]{79C78E} 1.9 & \cellcolor[HTML]{73C488} 1.8 & \cellcolor[HTML]{68C07F} 0.4 & \cellcolor[HTML]{63BE7B} 0.93 & \cellcolor[HTML]{89CD9B} 5.8 & \cellcolor[HTML]{AFDDBB} 2.7 & \cellcolor[HTML]{C8E8D0} 2.2 & \cellcolor[HTML]{6DC284} 1.9 & \cellcolor[HTML]{86CC99} 0.41 & \cellcolor[HTML]{6AC181} 0.95 & \textbf{0.22} \\
coral & \cellcolor[HTML]{64BE7C} 4 & \cellcolor[HTML]{82CB95} 1.8 & \cellcolor[HTML]{78C78D} 1.6 & \cellcolor[HTML]{63BE7B} 1.6 & \cellcolor[HTML]{D0EBD7} 0.27 & \cellcolor[HTML]{F4FAF5} 0.8 & \cellcolor[HTML]{90D0A1} 5 & \cellcolor[HTML]{A4D9B2} 2.5 & \cellcolor[HTML]{63BE7B} 1.9 & \cellcolor[HTML]{76C68B} 1.8 & \cellcolor[HTML]{63BE7B} 0.4 & \cellcolor[HTML]{97D3A7} 0.99 & \cellcolor[HTML]{73C488} 5.7 & \cellcolor[HTML]{73C488} 2.5 & \cellcolor[HTML]{80CA93} 2 & \cellcolor[HTML]{63BE7B} 1.9 & \cellcolor[HTML]{75C58A} 0.4 & \cellcolor[HTML]{63BE7B} 0.94 & \textbf{0.22} \\
xgb & \cellcolor[HTML]{65BF7D} 4 & \cellcolor[HTML]{7DC891} 1.8 & \cellcolor[HTML]{75C58A} 1.6 & \cellcolor[HTML]{6AC181} 1.6 & \cellcolor[HTML]{ADDDBA} 0.26 & \cellcolor[HTML]{63BE7B} 0.67 & \cellcolor[HTML]{63BE7B} 4.7 & \cellcolor[HTML]{71C386} 2.4 & \cellcolor[HTML]{99D4A9} 2 & \cellcolor[HTML]{63BE7B} 1.8 & \cellcolor[HTML]{90D1A1} 0.42 & \cellcolor[HTML]{FFFFFF} 1.1 & \cellcolor[HTML]{63BE7B} 5.6 & \cellcolor[HTML]{63BE7B} 2.5 & \cellcolor[HTML]{63BE7B} 1.9 & \cellcolor[HTML]{7EC992} 1.9 & \cellcolor[HTML]{D0EBD8} 0.45 & \cellcolor[HTML]{ECF7EF} 1.1 & \textbf{0.22} \\
gdro & \cellcolor[HTML]{63BE7B} 3.9 & \cellcolor[HTML]{76C68B} 1.8 & \cellcolor[HTML]{63BE7B} 1.6 & \cellcolor[HTML]{64BE7C} 1.6 & \cellcolor[HTML]{F5FAF6} 0.28 & \cellcolor[HTML]{EAF6ED} 0.79 & \cellcolor[HTML]{63BE7B} 4.7 & \cellcolor[HTML]{63BE7B} 2.3 & \cellcolor[HTML]{8BCE9D} 2 & \cellcolor[HTML]{78C78D} 1.9 & \cellcolor[HTML]{9ED6AD} 0.43 & \cellcolor[HTML]{FFFFFF} 1.2 & \cellcolor[HTML]{E1F2E6} 6.5 & \cellcolor[HTML]{FFFFFF} 3 & \cellcolor[HTML]{CDEAD4} 2.2 & \cellcolor[HTML]{7FCA93} 2 & \cellcolor[HTML]{63BE7B} 0.39 & \cellcolor[HTML]{FFFFFF} 1.7 & \textbf{0.17} \\
lr & \cellcolor[HTML]{FFFFFF} 6 & \cellcolor[HTML]{FFFFFF} 2.8 & \cellcolor[HTML]{FFFFFF} 2.6 & \cellcolor[HTML]{FFFFFF} 1.9 & \cellcolor[HTML]{93D2A4} 0.25 & \cellcolor[HTML]{FFFFFF} 1.2 & \cellcolor[HTML]{FFFFFF} 6.1 & \cellcolor[HTML]{FFFFFF} 2.9 & \cellcolor[HTML]{FFFFFF} 2.6 & \cellcolor[HTML]{D6EEDD} 2.1 & \cellcolor[HTML]{FFFFFF} 0.48 & \cellcolor[HTML]{FFFFFF} 1.4 & \cellcolor[HTML]{FFFFFF} 7.5 & \cellcolor[HTML]{FFFFFF} 3.1 & \cellcolor[HTML]{FFFFFF} 2.5 & \cellcolor[HTML]{FFFFFF} 2.3 & \cellcolor[HTML]{ADDDBA} 0.43 & \cellcolor[HTML]{FFFFFF} 1.5 & \textbf{0.00} \\
mmd & \cellcolor[HTML]{FFFFFF} 9.3 & \cellcolor[HTML]{FFFFFF} 5.6 & \cellcolor[HTML]{FFFFFF} 5.6 & \cellcolor[HTML]{FFFFFF} 2 & \cellcolor[HTML]{8DCF9E} 0.25 & \cellcolor[HTML]{FFFFFF} 3.4 & \cellcolor[HTML]{FFFFFF} 6.6 & \cellcolor[HTML]{FFFFFF} 3.1 & \cellcolor[HTML]{FFFFFF} 2.9 & \cellcolor[HTML]{EAF6ED} 2.1 & \cellcolor[HTML]{A4D9B2} 0.43 & \cellcolor[HTML]{FFFFFF} 1.6 & \cellcolor[HTML]{FFFFFF} 8.8 & \cellcolor[HTML]{FFFFFF} 4.3 & \cellcolor[HTML]{FFFFFF} 3.9 & \cellcolor[HTML]{E6F4EA} 2.2 & \cellcolor[HTML]{70C386} 0.4 & \cellcolor[HTML]{FFFFFF} 2.3 & \textbf{-0.35} \\
constant & \cellcolor[HTML]{FFFFFF} 9.3 & \cellcolor[HTML]{FFFFFF} 5.6 & \cellcolor[HTML]{FFFFFF} 5.5 & \cellcolor[HTML]{FFFFFF} 2 & \cellcolor[HTML]{8DCF9E} 0.25 & \cellcolor[HTML]{FFFFFF} 3.4 & \cellcolor[HTML]{FFFFFF} 9.5 & \cellcolor[HTML]{FFFFFF} 5.5 & \cellcolor[HTML]{FFFFFF} 5.4 & \cellcolor[HTML]{FFFFFF} 2.4 & \cellcolor[HTML]{F0F8F2} 0.47 & \cellcolor[HTML]{FFFFFF} 3 & \cellcolor[HTML]{FFFFFF} 11 & \cellcolor[HTML]{FFFFFF} 5.3 & \cellcolor[HTML]{FFFFFF} 5.1 & \cellcolor[HTML]{FAFDFB} 2.2 & \cellcolor[HTML]{DFF1E4} 0.46 & \cellcolor[HTML]{FFFFFF} 3 & \textbf{-0.64} \\
\bottomrule
\end{tabular}
}
    \label{tab:supp-et-median}
\end{table}

\begin{table}[htbp]
    \centering
    \caption{90th quantile of domain-level RMSE for ET $\downarrow$}
    {\small
\setlength{\tabcolsep}{1.4pt}
\begin{tabular}{lcccccc@{\hspace{1.5em}}cccccc@{\hspace{1.5em}}cccccc@{\hspace{1.5em}}c}
\toprule
 & \multicolumn{6}{c}{temporal} & \multicolumn{6}{c}{spatial} & \multicolumn{6}{c}{temperature} & \multirow{2}{*}{\rotatebox{90}{Skill score $\uparrow$}} \\
 & \rotatebox{90}{hourly} & \rotatebox{90}{weekly} & \rotatebox{90}{seasonal} & \rotatebox{90}{anom} & \rotatebox{90}{iav} & \rotatebox{90}{site-mean} & \rotatebox{90}{hourly} & \rotatebox{90}{weekly} & \rotatebox{90}{seasonal} & \rotatebox{90}{anom} & \rotatebox{90}{iav} & \rotatebox{90}{site-mean} & \rotatebox{90}{hourly} & \rotatebox{90}{weekly} & \rotatebox{90}{seasonal} & \rotatebox{90}{anom} & \rotatebox{90}{iav} & \rotatebox{90}{site-mean} &  \\
\midrule
coral & \cellcolor[HTML]{63BE7B} 6.4 & \cellcolor[HTML]{74C589} 3.6 & \cellcolor[HTML]{7BC890} 3 & \cellcolor[HTML]{64BE7C} 2.4 & \cellcolor[HTML]{88CD9B} 0.8 & \cellcolor[HTML]{7EC992} 2.2 & \cellcolor[HTML]{7BC88F} 7.6 & \cellcolor[HTML]{9BD5AB} 4 & \cellcolor[HTML]{AEDDBA} 3.6 & \cellcolor[HTML]{63BE7B} 2.6 & \cellcolor[HTML]{63BE7B} 0.84 & \cellcolor[HTML]{63BE7B} 2.7 & \cellcolor[HTML]{99D4A9} 9 & \cellcolor[HTML]{63BE7B} 3.9 & \cellcolor[HTML]{98D4A8} 3.4 & \cellcolor[HTML]{64BE7C} 2.5 & \cellcolor[HTML]{63BE7B} 0.82 & \cellcolor[HTML]{63BE7B} 2.7 & \textbf{0.17} \\
mlp & \cellcolor[HTML]{75C58A} 6.5 & \cellcolor[HTML]{74C589} 3.6 & \cellcolor[HTML]{63BE7B} 2.9 & \cellcolor[HTML]{7AC78E} 2.5 & \cellcolor[HTML]{A1D8B0} 0.82 & \cellcolor[HTML]{94D2A5} 2.3 & \cellcolor[HTML]{63BE7B} 7.3 & \cellcolor[HTML]{8ACE9C} 4 & \cellcolor[HTML]{A9DBB6} 3.5 & \cellcolor[HTML]{6BC182} 2.6 & \cellcolor[HTML]{64BE7C} 0.84 & \cellcolor[HTML]{63BE7B} 2.7 & \cellcolor[HTML]{98D4A8} 9 & \cellcolor[HTML]{74C589} 4 & \cellcolor[HTML]{63BE7B} 3.2 & \cellcolor[HTML]{63BE7B} 2.5 & \cellcolor[HTML]{8ED09F} 0.87 & \cellcolor[HTML]{78C78D} 2.8 & \textbf{0.17} \\
xgb & \cellcolor[HTML]{8FD0A0} 6.7 & \cellcolor[HTML]{63BE7B} 3.5 & \cellcolor[HTML]{65BE7C} 2.9 & \cellcolor[HTML]{64BE7C} 2.4 & \cellcolor[HTML]{CBE9D3} 0.86 & \cellcolor[HTML]{63BE7B} 2.1 & \cellcolor[HTML]{71C487} 7.5 & \cellcolor[HTML]{63BE7B} 3.8 & \cellcolor[HTML]{63BE7B} 3.3 & \cellcolor[HTML]{7CC890} 2.6 & \cellcolor[HTML]{65BE7C} 0.84 & \cellcolor[HTML]{7DC891} 2.8 & \cellcolor[HTML]{63BE7B} 8.4 & \cellcolor[HTML]{B2DFBE} 4.4 & \cellcolor[HTML]{AADBB7} 3.4 & \cellcolor[HTML]{71C487} 2.5 & \cellcolor[HTML]{94D2A4} 0.87 & \cellcolor[HTML]{D1EBD8} 3.1 & \textbf{0.16} \\
gdro & \cellcolor[HTML]{81CA94} 6.6 & \cellcolor[HTML]{69C080} 3.5 & \cellcolor[HTML]{79C78D} 3 & \cellcolor[HTML]{63BE7B} 2.4 & \cellcolor[HTML]{99D4A8} 0.81 & \cellcolor[HTML]{65BE7C} 2.1 & \cellcolor[HTML]{99D4A9} 7.8 & \cellcolor[HTML]{B1DEBD} 4.1 & \cellcolor[HTML]{E7F5EA} 3.8 & \cellcolor[HTML]{7CC890} 2.6 & \cellcolor[HTML]{6FC385} 0.85 & \cellcolor[HTML]{65BE7D} 2.7 & \cellcolor[HTML]{F4FAF6} 10 & \cellcolor[HTML]{FFFFFF} 7.3 & \cellcolor[HTML]{FFFFFF} 7 & \cellcolor[HTML]{97D3A7} 2.6 & \cellcolor[HTML]{A4D9B2} 0.89 & \cellcolor[HTML]{FFFFFF} 5.7 & \textbf{0.02} \\
lr & \cellcolor[HTML]{FFFFFF} 8.1 & \cellcolor[HTML]{FFFFFF} 4.3 & \cellcolor[HTML]{FFFFFF} 4.2 & \cellcolor[HTML]{9BD5AA} 2.6 & \cellcolor[HTML]{63BE7B} 0.76 & \cellcolor[HTML]{FFFFFF} 2.9 & \cellcolor[HTML]{EBF6EE} 8.6 & \cellcolor[HTML]{FFFFFF} 5.5 & \cellcolor[HTML]{FFFFFF} 5.4 & \cellcolor[HTML]{A3D9B1} 2.8 & \cellcolor[HTML]{C4E6CD} 0.94 & \cellcolor[HTML]{FFFFFF} 3.9 & \cellcolor[HTML]{FFFFFF} 10 & \cellcolor[HTML]{FFFFFF} 4.8 & \cellcolor[HTML]{FFFFFF} 4.3 & \cellcolor[HTML]{C9E8D1} 2.8 & \cellcolor[HTML]{9CD5AB} 0.88 & \cellcolor[HTML]{FFFFFF} 3.8 & \textbf{0.00} \\
mmd & \cellcolor[HTML]{FFFFFF} 13 & \cellcolor[HTML]{FFFFFF} 6.6 & \cellcolor[HTML]{FFFFFF} 6.6 & \cellcolor[HTML]{F9FCFA} 2.9 & \cellcolor[HTML]{A3D9B1} 0.82 & \cellcolor[HTML]{FFFFFF} 5.7 & \cellcolor[HTML]{FFFFFF} 12 & \cellcolor[HTML]{FFFFFF} 6.5 & \cellcolor[HTML]{FFFFFF} 5.2 & \cellcolor[HTML]{C2E5CB} 2.9 & \cellcolor[HTML]{64BE7C} 0.84 & \cellcolor[HTML]{FFFFFF} 3.9 & \cellcolor[HTML]{FFFFFF} 14 & \cellcolor[HTML]{FFFFFF} 5.7 & \cellcolor[HTML]{FFFFFF} 5.4 & \cellcolor[HTML]{F9FCFA} 3 & \cellcolor[HTML]{CEEAD5} 0.94 & \cellcolor[HTML]{FFFFFF} 4.3 & \textbf{-0.25} \\
constant & \cellcolor[HTML]{FFFFFF} 13 & \cellcolor[HTML]{FFFFFF} 6.5 & \cellcolor[HTML]{FFFFFF} 6.6 & \cellcolor[HTML]{F9FCFA} 2.9 & \cellcolor[HTML]{A3D9B1} 0.82 & \cellcolor[HTML]{FFFFFF} 5.6 & \cellcolor[HTML]{FFFFFF} 14 & \cellcolor[HTML]{FFFFFF} 6.9 & \cellcolor[HTML]{FFFFFF} 7 & \cellcolor[HTML]{FFFFFF} 3.1 & \cellcolor[HTML]{A9DBB7} 0.92 & \cellcolor[HTML]{FFFFFF} 5.4 & \cellcolor[HTML]{FFFFFF} 16 & \cellcolor[HTML]{FFFFFF} 6.6 & \cellcolor[HTML]{FFFFFF} 6 & \cellcolor[HTML]{FFFFFF} 3.2 & \cellcolor[HTML]{FFFFFF} 1.1 & \cellcolor[HTML]{FFFFFF} 5.4 & \textbf{-0.37} \\
\bottomrule
\end{tabular}
}
    \label{tab:supp-et-90}
\end{table}

\begin{table}[htbp]
    \centering
    \caption{Median of domain-level RMSE for GPP $\downarrow$}
    {\small
\setlength{\tabcolsep}{1.4pt}
\begin{tabular}{lcccccc@{\hspace{1.5em}}cccccc@{\hspace{1.5em}}cccccc@{\hspace{1.5em}}c}
\toprule
 & \multicolumn{6}{c}{temporal} & \multicolumn{6}{c}{spatial} & \multicolumn{6}{c}{temperature} & \multirow{2}{*}{\rotatebox{90}{Skill score $\uparrow$}} \\
 & \rotatebox{90}{hourly} & \rotatebox{90}{weekly} & \rotatebox{90}{seasonal} & \rotatebox{90}{anom} & \rotatebox{90}{iav} & \rotatebox{90}{site-mean} & \rotatebox{90}{hourly} & \rotatebox{90}{weekly} & \rotatebox{90}{seasonal} & \rotatebox{90}{anom} & \rotatebox{90}{iav} & \rotatebox{90}{site-mean} & \rotatebox{90}{hourly} & \rotatebox{90}{weekly} & \rotatebox{90}{seasonal} & \rotatebox{90}{anom} & \rotatebox{90}{iav} & \rotatebox{90}{site-mean} &  \\
\midrule
mmd & \cellcolor[HTML]{63BE7B} 2.3 & \cellcolor[HTML]{63BE7B} 1.1 & \cellcolor[HTML]{63BE7B} 0.95 & \cellcolor[HTML]{6AC181} 0.9 & \cellcolor[HTML]{63BE7B} 0.18 & \cellcolor[HTML]{74C58A} 0.42 & \cellcolor[HTML]{63BE7B} 3.2 & \cellcolor[HTML]{63BE7B} 1.7 & \cellcolor[HTML]{63BE7B} 1.2 & \cellcolor[HTML]{63BE7B} 1.2 & \cellcolor[HTML]{63BE7B} 0.33 & \cellcolor[HTML]{E6F4E9} 0.57 & \cellcolor[HTML]{8ED0A0} 3.2 & \cellcolor[HTML]{64BE7B} 1.3 & \cellcolor[HTML]{63BE7B} 1 & \cellcolor[HTML]{65BE7D} 0.83 & \cellcolor[HTML]{BAE2C5} 0.2 & \cellcolor[HTML]{BAE2C4} 0.55 & \textbf{0.29} \\
mlp & \cellcolor[HTML]{63BE7B} 2.3 & \cellcolor[HTML]{71C386} 1.1 & \cellcolor[HTML]{84CC97} 1 & \cellcolor[HTML]{64BE7C} 0.89 & \cellcolor[HTML]{ABDCB7} 0.2 & \cellcolor[HTML]{8ACE9C} 0.43 & \cellcolor[HTML]{97D3A7} 3.5 & \cellcolor[HTML]{B5E0C1} 1.9 & \cellcolor[HTML]{C9E8D1} 1.3 & \cellcolor[HTML]{7FC993} 1.2 & \cellcolor[HTML]{74C58A} 0.33 & \cellcolor[HTML]{63BE7B} 0.49 & \cellcolor[HTML]{84CC97} 3.1 & \cellcolor[HTML]{63BE7B} 1.3 & \cellcolor[HTML]{6EC284} 1 & \cellcolor[HTML]{65BF7D} 0.83 & \cellcolor[HTML]{B4DFBF} 0.2 & \cellcolor[HTML]{AFDDBB} 0.54 & \textbf{0.28} \\
coral & \cellcolor[HTML]{6FC385} 2.4 & \cellcolor[HTML]{72C487} 1.1 & \cellcolor[HTML]{9FD7AD} 1 & \cellcolor[HTML]{72C488} 0.91 & \cellcolor[HTML]{96D3A6} 0.19 & \cellcolor[HTML]{DEF1E3} 0.48 & \cellcolor[HTML]{64BE7C} 3.2 & \cellcolor[HTML]{6CC182} 1.8 & \cellcolor[HTML]{6DC284} 1.2 & \cellcolor[HTML]{64BE7C} 1.2 & \cellcolor[HTML]{63BE7B} 0.33 & \cellcolor[HTML]{FFFFFF} 0.59 & \cellcolor[HTML]{85CC98} 3.1 & \cellcolor[HTML]{88CD9B} 1.3 & \cellcolor[HTML]{D1EBD8} 1.1 & \cellcolor[HTML]{66BF7D} 0.83 & \cellcolor[HTML]{63BE7B} 0.18 & \cellcolor[HTML]{FFFFFF} 0.62 & \textbf{0.27} \\
xgb & \cellcolor[HTML]{73C488} 2.4 & \cellcolor[HTML]{7AC78E} 1.2 & \cellcolor[HTML]{BDE3C7} 1.1 & \cellcolor[HTML]{63BE7B} 0.89 & \cellcolor[HTML]{F5FAF6} 0.22 & \cellcolor[HTML]{63BE7B} 0.41 & \cellcolor[HTML]{9AD5A9} 3.5 & \cellcolor[HTML]{BCE3C6} 1.9 & \cellcolor[HTML]{B8E1C3} 1.3 & \cellcolor[HTML]{69C080} 1.2 & \cellcolor[HTML]{8BCE9C} 0.34 & \cellcolor[HTML]{FFFFFF} 0.67 & \cellcolor[HTML]{63BE7B} 3 & \cellcolor[HTML]{6AC181} 1.3 & \cellcolor[HTML]{77C68C} 1 & \cellcolor[HTML]{63BE7B} 0.83 & \cellcolor[HTML]{EDF7F0} 0.21 & \cellcolor[HTML]{63BE7B} 0.49 & \textbf{0.26} \\
lr & \cellcolor[HTML]{FFFFFF} 4.3 & \cellcolor[HTML]{FFFFFF} 2.1 & \cellcolor[HTML]{FFFFFF} 1.9 & \cellcolor[HTML]{FFFFFF} 1.1 & \cellcolor[HTML]{C7E7CF} 0.2 & \cellcolor[HTML]{FFFFFF} 0.89 & \cellcolor[HTML]{FFFFFF} 4.5 & \cellcolor[HTML]{FFFFFF} 2.1 & \cellcolor[HTML]{FFFFFF} 1.7 & \cellcolor[HTML]{F8FCF9} 1.4 & \cellcolor[HTML]{FFFFFF} 0.4 & \cellcolor[HTML]{FFFFFF} 0.82 & \cellcolor[HTML]{FFFFFF} 4.6 & \cellcolor[HTML]{FFFFFF} 1.9 & \cellcolor[HTML]{FFFFFF} 1.5 & \cellcolor[HTML]{81CA94} 0.86 & \cellcolor[HTML]{B0DEBC} 0.2 & \cellcolor[HTML]{FFFFFF} 1.2 & \textbf{0.00} \\
gdro & \cellcolor[HTML]{FFFFFF} 3.4 & \cellcolor[HTML]{FFFFFF} 1.8 & \cellcolor[HTML]{FFFFFF} 1.7 & \cellcolor[HTML]{C7E7D0} 1 & \cellcolor[HTML]{DBF0E1} 0.21 & \cellcolor[HTML]{FFFFFF} 0.75 & \cellcolor[HTML]{C8E8D1} 3.7 & \cellcolor[HTML]{9FD7AE} 1.9 & \cellcolor[HTML]{B6E0C1} 1.3 & \cellcolor[HTML]{C3E6CC} 1.3 & \cellcolor[HTML]{FFFFFF} 0.41 & \cellcolor[HTML]{A0D7AF} 0.53 & \cellcolor[HTML]{FFFFFF} 4.2 & \cellcolor[HTML]{FFFFFF} 2.7 & \cellcolor[HTML]{FFFFFF} 2.5 & \cellcolor[HTML]{FFFFFF} 1 & \cellcolor[HTML]{FFFFFF} 0.25 & \cellcolor[HTML]{FFFFFF} 2 & \textbf{-0.02} \\
constant & \cellcolor[HTML]{FFFFFF} 6.2 & \cellcolor[HTML]{FFFFFF} 3.9 & \cellcolor[HTML]{FFFFFF} 3.8 & \cellcolor[HTML]{FFFFFF} 1.1 & \cellcolor[HTML]{F4FAF6} 0.22 & \cellcolor[HTML]{FFFFFF} 1.7 & \cellcolor[HTML]{FFFFFF} 6.9 & \cellcolor[HTML]{FFFFFF} 3.8 & \cellcolor[HTML]{FFFFFF} 3.5 & \cellcolor[HTML]{F8FCF9} 1.4 & \cellcolor[HTML]{ECF7EF} 0.38 & \cellcolor[HTML]{FFFFFF} 1.1 & \cellcolor[HTML]{FFFFFF} 5.4 & \cellcolor[HTML]{FFFFFF} 3.3 & \cellcolor[HTML]{FFFFFF} 3.1 & \cellcolor[HTML]{FFFFFF} 1 & \cellcolor[HTML]{FFFFFF} 0.28 & \cellcolor[HTML]{FFFFFF} 1.7 & \textbf{-0.50} \\
\bottomrule
\end{tabular}
}
    \label{tab:supp-gpp-median}
\end{table}

\begin{table}[htbp]
    \centering
    \caption{90th quantile of domain-level RMSE for GPP $\downarrow$}
    {\small
\setlength{\tabcolsep}{1.4pt}
\begin{tabular}{lcccccc@{\hspace{1.5em}}cccccc@{\hspace{1.5em}}cccccc@{\hspace{1.5em}}c}
\toprule
 & \multicolumn{6}{c}{temporal} & \multicolumn{6}{c}{spatial} & \multicolumn{6}{c}{temperature} & \multirow{2}{*}{\rotatebox{90}{Skill score $\uparrow$}} \\
 & \rotatebox{90}{hourly} & \rotatebox{90}{weekly} & \rotatebox{90}{seasonal} & \rotatebox{90}{anom} & \rotatebox{90}{iav} & \rotatebox{90}{site-mean} & \rotatebox{90}{hourly} & \rotatebox{90}{weekly} & \rotatebox{90}{seasonal} & \rotatebox{90}{anom} & \rotatebox{90}{iav} & \rotatebox{90}{site-mean} & \rotatebox{90}{hourly} & \rotatebox{90}{weekly} & \rotatebox{90}{seasonal} & \rotatebox{90}{anom} & \rotatebox{90}{iav} & \rotatebox{90}{site-mean} &  \\
\midrule
coral & \cellcolor[HTML]{74C589} 5.2 & \cellcolor[HTML]{7CC890} 3 & \cellcolor[HTML]{63BE7B} 2.3 & \cellcolor[HTML]{7AC78E} 2.2 & \cellcolor[HTML]{63BE7B} 0.63 & \cellcolor[HTML]{63BE7B} 1.3 & \cellcolor[HTML]{68C07F} 4.8 & \cellcolor[HTML]{63BE7B} 2.9 & \cellcolor[HTML]{63BE7B} 2.4 & \cellcolor[HTML]{65BE7C} 2 & \cellcolor[HTML]{92D1A3} 0.79 & \cellcolor[HTML]{81CA94} 1.8 & \cellcolor[HTML]{95D2A5} 5.2 & \cellcolor[HTML]{95D3A6} 3 & \cellcolor[HTML]{B0DEBC} 2.5 & \cellcolor[HTML]{72C488} 1.5 & \cellcolor[HTML]{92D1A3} 0.56 & \cellcolor[HTML]{EDF7EF} 2 & \textbf{0.17} \\
xgb & \cellcolor[HTML]{6FC385} 5.2 & \cellcolor[HTML]{66BF7D} 2.9 & \cellcolor[HTML]{66BF7E} 2.3 & \cellcolor[HTML]{63BE7B} 2.2 & \cellcolor[HTML]{7AC78F} 0.64 & \cellcolor[HTML]{6EC285} 1.3 & \cellcolor[HTML]{C6E7CF} 5.3 & \cellcolor[HTML]{D9EFDE} 3.4 & \cellcolor[HTML]{B7E1C2} 2.6 & \cellcolor[HTML]{65BE7C} 2 & \cellcolor[HTML]{B0DEBC} 0.82 & \cellcolor[HTML]{93D2A4} 1.8 & \cellcolor[HTML]{63BE7B} 4.9 & \cellcolor[HTML]{63BE7B} 2.9 & \cellcolor[HTML]{63BE7B} 2.3 & \cellcolor[HTML]{63BE7B} 1.4 & \cellcolor[HTML]{A5D9B2} 0.57 & \cellcolor[HTML]{63BE7B} 1.7 & \textbf{0.17} \\
mmd & \cellcolor[HTML]{67BF7E} 5.1 & \cellcolor[HTML]{70C386} 2.9 & \cellcolor[HTML]{63BE7B} 2.3 & \cellcolor[HTML]{6EC284} 2.2 & \cellcolor[HTML]{64BE7C} 0.63 & \cellcolor[HTML]{69C080} 1.3 & \cellcolor[HTML]{63BE7B} 4.7 & \cellcolor[HTML]{81CA94} 3 & \cellcolor[HTML]{B7E1C2} 2.6 & \cellcolor[HTML]{63BE7B} 2 & \cellcolor[HTML]{90D0A1} 0.79 & \cellcolor[HTML]{A4D9B2} 1.9 & \cellcolor[HTML]{9ED6AD} 5.3 & \cellcolor[HTML]{C1E5CA} 3.2 & \cellcolor[HTML]{E5F4E9} 2.7 & \cellcolor[HTML]{6BC182} 1.5 & \cellcolor[HTML]{63BE7B} 0.52 & \cellcolor[HTML]{FFFFFF} 2 & \textbf{0.16} \\
mlp & \cellcolor[HTML]{63BE7B} 5.1 & \cellcolor[HTML]{63BE7B} 2.9 & \cellcolor[HTML]{64BE7C} 2.3 & \cellcolor[HTML]{63BE7B} 2.2 & \cellcolor[HTML]{92D1A3} 0.66 & \cellcolor[HTML]{A6DAB3} 1.4 & \cellcolor[HTML]{95D2A5} 5 & \cellcolor[HTML]{A1D7AF} 3.1 & \cellcolor[HTML]{E0F2E5} 2.8 & \cellcolor[HTML]{67BF7E} 2 & \cellcolor[HTML]{AFDDBB} 0.82 & \cellcolor[HTML]{89CE9B} 1.8 & \cellcolor[HTML]{9ED6AD} 5.3 & \cellcolor[HTML]{BFE4C9} 3.2 & \cellcolor[HTML]{D2ECD9} 2.6 & \cellcolor[HTML]{6EC284} 1.5 & \cellcolor[HTML]{63BE7B} 0.52 & \cellcolor[HTML]{F0F8F2} 2 & \textbf{0.15} \\
lr & \cellcolor[HTML]{FFFFFF} 6.3 & \cellcolor[HTML]{F9FCFA} 3.5 & \cellcolor[HTML]{FFFFFF} 2.9 & \cellcolor[HTML]{71C386} 2.2 & \cellcolor[HTML]{B1DEBD} 0.69 & \cellcolor[HTML]{FFFFFF} 2.5 & \cellcolor[HTML]{FFFFFF} 6 & \cellcolor[HTML]{FFFFFF} 3.5 & \cellcolor[HTML]{FFFFFF} 3 & \cellcolor[HTML]{CCEAD4} 2.2 & \cellcolor[HTML]{E5F4E9} 0.87 & \cellcolor[HTML]{FFFFFF} 2.4 & \cellcolor[HTML]{FFFFFF} 6.1 & \cellcolor[HTML]{FFFFFF} 3.6 & \cellcolor[HTML]{FFFFFF} 3.3 & \cellcolor[HTML]{80CA94} 1.5 & \cellcolor[HTML]{D6EEDC} 0.6 & \cellcolor[HTML]{FFFFFF} 2.8 & \textbf{0.00} \\
gdro & \cellcolor[HTML]{A5D9B3} 5.6 & \cellcolor[HTML]{F0F8F2} 3.4 & \cellcolor[HTML]{FFFFFF} 2.9 & \cellcolor[HTML]{A5D9B3} 2.4 & \cellcolor[HTML]{9ED6AD} 0.67 & \cellcolor[HTML]{FFFFFF} 1.9 & \cellcolor[HTML]{89CD9B} 5 & \cellcolor[HTML]{6AC181} 2.9 & \cellcolor[HTML]{99D4A9} 2.5 & \cellcolor[HTML]{80CA93} 2 & \cellcolor[HTML]{FFFFFF} 0.94 & \cellcolor[HTML]{63BE7B} 1.7 & \cellcolor[HTML]{FFFFFF} 6.9 & \cellcolor[HTML]{FFFFFF} 4.9 & \cellcolor[HTML]{FFFFFF} 4.8 & \cellcolor[HTML]{E0F2E4} 1.7 & \cellcolor[HTML]{B6E0C1} 0.58 & \cellcolor[HTML]{FFFFFF} 4.2 & \textbf{-0.02} \\
constant & \cellcolor[HTML]{FFFFFF} 9.3 & \cellcolor[HTML]{FFFFFF} 5 & \cellcolor[HTML]{FFFFFF} 4.7 & \cellcolor[HTML]{C4E6CD} 2.5 & \cellcolor[HTML]{DBF0E1} 0.72 & \cellcolor[HTML]{FFFFFF} 4 & \cellcolor[HTML]{FFFFFF} 9.2 & \cellcolor[HTML]{FFFFFF} 4.5 & \cellcolor[HTML]{FFFFFF} 4.2 & \cellcolor[HTML]{F7FBF8} 2.3 & \cellcolor[HTML]{63BE7B} 0.75 & \cellcolor[HTML]{FFFFFF} 3.1 & \cellcolor[HTML]{FFFFFF} 8.5 & \cellcolor[HTML]{FFFFFF} 4.7 & \cellcolor[HTML]{FFFFFF} 4.3 & \cellcolor[HTML]{C7E7CF} 1.6 & \cellcolor[HTML]{FFFFFF} 0.73 & \cellcolor[HTML]{FFFFFF} 4.1 & \textbf{-0.30} \\
\bottomrule
\end{tabular}
}
    \label{tab:sup-gpp-90}
\end{table}

\begin{table}[htbp]
    \centering
    \caption{Median of domain-level RMSE for NEE $\downarrow$}
    {\small
\setlength{\tabcolsep}{1.4pt}
\begin{tabular}{lcccccc@{\hspace{1.5em}}cccccc@{\hspace{1.5em}}cccccc@{\hspace{1.5em}}c}
\toprule
 & \multicolumn{6}{c}{temporal} & \multicolumn{6}{c}{spatial} & \multicolumn{6}{c}{temperature} & \multirow{2}{*}{\rotatebox{90}{Skill score $\uparrow$}} \\
 & \rotatebox{90}{hourly} & \rotatebox{90}{weekly} & \rotatebox{90}{seasonal} & \rotatebox{90}{anom} & \rotatebox{90}{iav} & \rotatebox{90}{site-mean} & \rotatebox{90}{hourly} & \rotatebox{90}{weekly} & \rotatebox{90}{seasonal} & \rotatebox{90}{anom} & \rotatebox{90}{iav} & \rotatebox{90}{site-mean} & \rotatebox{90}{hourly} & \rotatebox{90}{weekly} & \rotatebox{90}{seasonal} & \rotatebox{90}{anom} & \rotatebox{90}{iav} & \rotatebox{90}{site-mean} &  \\
\midrule
mmd & \cellcolor[HTML]{63BE7B} 2.2 & \cellcolor[HTML]{63BE7B} 0.93 & \cellcolor[HTML]{6EC284} 0.76 & \cellcolor[HTML]{6CC182} 0.75 & \cellcolor[HTML]{D1ECD8} 0.13 & \cellcolor[HTML]{6CC182} 0.3 & \cellcolor[HTML]{67C07F} 2.9 & \cellcolor[HTML]{68C07F} 1.4 & \cellcolor[HTML]{63BE7B} 0.94 & \cellcolor[HTML]{63BE7B} 0.98 & \cellcolor[HTML]{7BC88F} 0.19 & \cellcolor[HTML]{63BE7B} 0.3 & \cellcolor[HTML]{63BE7B} 2.9 & \cellcolor[HTML]{B5E0C0} 1.1 & \cellcolor[HTML]{F9FCFA} 0.93 & \cellcolor[HTML]{6BC182} 0.73 & \cellcolor[HTML]{CBE9D3} 0.14 & \cellcolor[HTML]{63BE7B} 0.48 & \textbf{0.33} \\
coral & \cellcolor[HTML]{63BE7B} 2.2 & \cellcolor[HTML]{70C386} 0.95 & \cellcolor[HTML]{81CA94} 0.78 & \cellcolor[HTML]{6DC284} 0.75 & \cellcolor[HTML]{C7E7D0} 0.13 & \cellcolor[HTML]{63BE7B} 0.29 & \cellcolor[HTML]{63BE7B} 2.9 & \cellcolor[HTML]{63BE7B} 1.4 & \cellcolor[HTML]{84CB97} 0.98 & \cellcolor[HTML]{63BE7B} 0.98 & \cellcolor[HTML]{75C58A} 0.19 & \cellcolor[HTML]{FFFFFF} 0.38 & \cellcolor[HTML]{70C386} 2.9 & \cellcolor[HTML]{63BE7B} 1 & \cellcolor[HTML]{CFEBD7} 0.89 & \cellcolor[HTML]{63BE7B} 0.72 & \cellcolor[HTML]{C6E7CF} 0.14 & \cellcolor[HTML]{A1D7AF} 0.52 & \textbf{0.33} \\
mlp & \cellcolor[HTML]{6BC182} 2.3 & \cellcolor[HTML]{6FC385} 0.95 & \cellcolor[HTML]{63BE7B} 0.75 & \cellcolor[HTML]{6FC385} 0.75 & \cellcolor[HTML]{6FC385} 0.11 & \cellcolor[HTML]{74C589} 0.3 & \cellcolor[HTML]{6CC283} 2.9 & \cellcolor[HTML]{6EC284} 1.4 & \cellcolor[HTML]{91D1A2} 1 & \cellcolor[HTML]{89CE9B} 1 & \cellcolor[HTML]{6AC181} 0.19 & \cellcolor[HTML]{FFFFFF} 0.38 & \cellcolor[HTML]{73C589} 2.9 & \cellcolor[HTML]{B7E1C2} 1.1 & \cellcolor[HTML]{FFFFFF} 0.96 & \cellcolor[HTML]{6EC285} 0.73 & \cellcolor[HTML]{D6EEDC} 0.14 & \cellcolor[HTML]{DFF1E4} 0.55 & \textbf{0.32} \\
xgb & \cellcolor[HTML]{70C386} 2.3 & \cellcolor[HTML]{7BC88F} 0.96 & \cellcolor[HTML]{9FD7AE} 0.81 & \cellcolor[HTML]{63BE7B} 0.74 & \cellcolor[HTML]{63BE7B} 0.11 & \cellcolor[HTML]{77C68C} 0.3 & \cellcolor[HTML]{83CB96} 3 & \cellcolor[HTML]{84CC97} 1.5 & \cellcolor[HTML]{91D1A2} 1 & \cellcolor[HTML]{6EC284} 0.99 & \cellcolor[HTML]{6EC284} 0.19 & \cellcolor[HTML]{FFFFFF} 0.43 & \cellcolor[HTML]{68C07F} 2.9 & \cellcolor[HTML]{E8F5EB} 1.2 & \cellcolor[HTML]{63BE7B} 0.78 & \cellcolor[HTML]{83CB96} 0.75 & \cellcolor[HTML]{D3ECDA} 0.14 & \cellcolor[HTML]{F3FAF5} 0.57 & \textbf{0.31} \\
lr & \cellcolor[HTML]{FFFFFF} 4 & \cellcolor[HTML]{FFFFFF} 1.7 & \cellcolor[HTML]{FFFFFF} 1.6 & \cellcolor[HTML]{FFFFFF} 0.93 & \cellcolor[HTML]{FFFFFF} 0.16 & \cellcolor[HTML]{FFFFFF} 0.9 & \cellcolor[HTML]{FFFFFF} 4.1 & \cellcolor[HTML]{E0F2E4} 1.6 & \cellcolor[HTML]{FFFFFF} 1.4 & \cellcolor[HTML]{FFFFFF} 1.2 & \cellcolor[HTML]{FBFDFB} 0.22 & \cellcolor[HTML]{FFFFFF} 0.68 & \cellcolor[HTML]{FFFFFF} 4.5 & \cellcolor[HTML]{FFFFFF} 1.8 & \cellcolor[HTML]{FFFFFF} 1.7 & \cellcolor[HTML]{C4E6CD} 0.81 & \cellcolor[HTML]{63BE7B} 0.13 & \cellcolor[HTML]{FFFFFF} 1.4 & \textbf{0.00} \\
gdro & \cellcolor[HTML]{FFFFFF} 3.5 & \cellcolor[HTML]{FFFFFF} 1.9 & \cellcolor[HTML]{FFFFFF} 1.7 & \cellcolor[HTML]{E1F2E5} 0.86 & \cellcolor[HTML]{FFFFFF} 0.14 & \cellcolor[HTML]{FFFFFF} 0.89 & \cellcolor[HTML]{FFFFFF} 3.6 & \cellcolor[HTML]{FFFFFF} 1.8 & \cellcolor[HTML]{FFFFFF} 1.5 & \cellcolor[HTML]{FDFEFD} 1.2 & \cellcolor[HTML]{FFFFFF} 0.23 & \cellcolor[HTML]{FFFFFF} 0.85 & \cellcolor[HTML]{FFFFFF} 3.7 & \cellcolor[HTML]{FFFFFF} 2.3 & \cellcolor[HTML]{FFFFFF} 2.2 & \cellcolor[HTML]{FFFFFF} 0.89 & \cellcolor[HTML]{76C68B} 0.13 & \cellcolor[HTML]{FFFFFF} 1.7 & \textbf{-0.05} \\
constant & \cellcolor[HTML]{FFFFFF} 5.4 & \cellcolor[HTML]{FFFFFF} 2.2 & \cellcolor[HTML]{FFFFFF} 2.2 & \cellcolor[HTML]{FFFFFF} 0.93 & \cellcolor[HTML]{EDF7F0} 0.13 & \cellcolor[HTML]{FFFFFF} 1.7 & \cellcolor[HTML]{FFFFFF} 6.3 & \cellcolor[HTML]{FFFFFF} 2.3 & \cellcolor[HTML]{FFFFFF} 2.1 & \cellcolor[HTML]{FFFFFF} 1.3 & \cellcolor[HTML]{63BE7B} 0.18 & \cellcolor[HTML]{FFFFFF} 1.5 & \cellcolor[HTML]{FFFFFF} 5 & \cellcolor[HTML]{FFFFFF} 2 & \cellcolor[HTML]{FFFFFF} 2 & \cellcolor[HTML]{81CA95} 0.75 & \cellcolor[HTML]{FFFFFF} 0.17 & \cellcolor[HTML]{FFFFFF} 1.7 & \textbf{-0.29} \\
\bottomrule
\end{tabular}
}
    \label{tab:nee-supp-med}
\end{table}

\begin{table}[htbp]
    \centering
    \caption{90th quantile of domain-level RMSE for NEE $\downarrow$}
    {\small
\setlength{\tabcolsep}{1.4pt}
\begin{tabular}{lcccccc@{\hspace{1.5em}}cccccc@{\hspace{1.5em}}cccccc@{\hspace{1.5em}}c}
\toprule
 & \multicolumn{6}{c}{temporal} & \multicolumn{6}{c}{spatial} & \multicolumn{6}{c}{temperature} & \multirow{2}{*}{\rotatebox{90}{Skill score $\uparrow$}} \\
 & \rotatebox{90}{hourly} & \rotatebox{90}{weekly} & \rotatebox{90}{seasonal} & \rotatebox{90}{anom} & \rotatebox{90}{iav} & \rotatebox{90}{site-mean} & \rotatebox{90}{hourly} & \rotatebox{90}{weekly} & \rotatebox{90}{seasonal} & \rotatebox{90}{anom} & \rotatebox{90}{iav} & \rotatebox{90}{site-mean} & \rotatebox{90}{hourly} & \rotatebox{90}{weekly} & \rotatebox{90}{seasonal} & \rotatebox{90}{anom} & \rotatebox{90}{iav} & \rotatebox{90}{site-mean} &  \\
\midrule
xgb & \cellcolor[HTML]{63BE7B} 4.7 & \cellcolor[HTML]{68C07F} 2.4 & \cellcolor[HTML]{63BE7B} 1.7 & \cellcolor[HTML]{67BF7E} 1.7 & \cellcolor[HTML]{63BE7B} 0.4 & \cellcolor[HTML]{63BE7B} 1 & \cellcolor[HTML]{64BE7C} 4.5 & \cellcolor[HTML]{7CC890} 2.4 & \cellcolor[HTML]{D9EFDE} 2.1 & \cellcolor[HTML]{63BE7B} 1.7 & \cellcolor[HTML]{6AC181} 0.58 & \cellcolor[HTML]{8ACE9C} 1.3 & \cellcolor[HTML]{78C78D} 4.9 & \cellcolor[HTML]{D8EFDE} 2 & \cellcolor[HTML]{63BE7B} 1.5 & \cellcolor[HTML]{83CB96} 1.2 & \cellcolor[HTML]{63BE7B} 0.33 & \cellcolor[HTML]{63BE7B} 1.2 & \textbf{0.24} \\
coral & \cellcolor[HTML]{79C78E} 4.9 & \cellcolor[HTML]{72C488} 2.4 & \cellcolor[HTML]{CEEAD5} 1.9 & \cellcolor[HTML]{65BF7D} 1.7 & \cellcolor[HTML]{6FC385} 0.41 & \cellcolor[HTML]{70C386} 1.1 & \cellcolor[HTML]{8ACE9C} 4.7 & \cellcolor[HTML]{89CD9B} 2.4 & \cellcolor[HTML]{7CC890} 1.9 & \cellcolor[HTML]{64BE7C} 1.7 & \cellcolor[HTML]{71C487} 0.58 & \cellcolor[HTML]{7DC991} 1.3 & \cellcolor[HTML]{63BE7B} 4.8 & \cellcolor[HTML]{63BE7B} 1.7 & \cellcolor[HTML]{68C07F} 1.5 & \cellcolor[HTML]{66BF7D} 1.2 & \cellcolor[HTML]{73C488} 0.33 & \cellcolor[HTML]{B6E0C1} 1.3 & \textbf{0.23} \\
mmd & \cellcolor[HTML]{71C487} 4.8 & \cellcolor[HTML]{63BE7B} 2.4 & \cellcolor[HTML]{9BD5AA} 1.8 & \cellcolor[HTML]{74C589} 1.8 & \cellcolor[HTML]{75C58B} 0.41 & \cellcolor[HTML]{97D4A7} 1.1 & \cellcolor[HTML]{63BE7B} 4.5 & \cellcolor[HTML]{63BE7B} 2.3 & \cellcolor[HTML]{63BE7B} 1.9 & \cellcolor[HTML]{64BE7C} 1.7 & \cellcolor[HTML]{66BF7D} 0.58 & \cellcolor[HTML]{63BE7B} 1.3 & \cellcolor[HTML]{63BE7B} 4.8 & \cellcolor[HTML]{E7F5EB} 2 & \cellcolor[HTML]{AADBB7} 1.6 & \cellcolor[HTML]{63BE7B} 1.2 & \cellcolor[HTML]{68C07F} 0.33 & \cellcolor[HTML]{EEF8F0} 1.4 & \textbf{0.23} \\
mlp & \cellcolor[HTML]{72C487} 4.8 & \cellcolor[HTML]{7FCA93} 2.5 & \cellcolor[HTML]{C8E8D0} 1.9 & \cellcolor[HTML]{63BE7B} 1.7 & \cellcolor[HTML]{64BE7C} 0.4 & \cellcolor[HTML]{A1D8AF} 1.1 & \cellcolor[HTML]{87CD99} 4.7 & \cellcolor[HTML]{8FD0A0} 2.5 & \cellcolor[HTML]{A9DBB6} 2 & \cellcolor[HTML]{6CC283} 1.7 & \cellcolor[HTML]{63BE7B} 0.57 & \cellcolor[HTML]{FAFDFB} 1.5 & \cellcolor[HTML]{68C07F} 4.8 & \cellcolor[HTML]{F2F9F4} 2 & \cellcolor[HTML]{CEEAD6} 1.7 & \cellcolor[HTML]{64BE7C} 1.2 & \cellcolor[HTML]{66BF7D} 0.33 & \cellcolor[HTML]{BCE3C6} 1.3 & \textbf{0.22} \\
lr & \cellcolor[HTML]{FFFFFF} 5.9 & \cellcolor[HTML]{FFFFFF} 3 & \cellcolor[HTML]{FFFFFF} 2.5 & \cellcolor[HTML]{A0D7AF} 1.9 & \cellcolor[HTML]{82CB95} 0.42 & \cellcolor[HTML]{FFFFFF} 2.1 & \cellcolor[HTML]{FFFFFF} 5.8 & \cellcolor[HTML]{FFFFFF} 3.1 & \cellcolor[HTML]{FFFFFF} 3 & \cellcolor[HTML]{9AD5A9} 1.8 & \cellcolor[HTML]{8ACE9C} 0.6 & \cellcolor[HTML]{FFFFFF} 2.3 & \cellcolor[HTML]{EFF8F2} 5.6 & \cellcolor[HTML]{FFFFFF} 3 & \cellcolor[HTML]{FFFFFF} 3 & \cellcolor[HTML]{B6E0C1} 1.3 & \cellcolor[HTML]{D1ECD8} 0.37 & \cellcolor[HTML]{FFFFFF} 2.5 & \textbf{0.00} \\
gdro & \cellcolor[HTML]{CAE9D2} 5.4 & \cellcolor[HTML]{FFFFFF} 3.2 & \cellcolor[HTML]{FFFFFF} 2.9 & \cellcolor[HTML]{85CC98} 1.8 & \cellcolor[HTML]{B2DEBD} 0.44 & \cellcolor[HTML]{FFFFFF} 2.3 & \cellcolor[HTML]{9FD7AE} 4.9 & \cellcolor[HTML]{FFFFFF} 3.1 & \cellcolor[HTML]{FFFFFF} 2.7 & \cellcolor[HTML]{7BC88F} 1.7 & \cellcolor[HTML]{63BE7B} 0.57 & \cellcolor[HTML]{FFFFFF} 2.6 & \cellcolor[HTML]{E8F5EB} 5.6 & \cellcolor[HTML]{FFFFFF} 3.4 & \cellcolor[HTML]{FFFFFF} 3.3 & \cellcolor[HTML]{96D3A6} 1.3 & \cellcolor[HTML]{D1EBD8} 0.37 & \cellcolor[HTML]{FFFFFF} 3 & \textbf{-0.02} \\
constant & \cellcolor[HTML]{FFFFFF} 8.1 & \cellcolor[HTML]{FFFFFF} 3.4 & \cellcolor[HTML]{FFFFFF} 3.2 & \cellcolor[HTML]{FFFFFF} 2.1 & \cellcolor[HTML]{CDEAD5} 0.46 & \cellcolor[HTML]{FFFFFF} 2.2 & \cellcolor[HTML]{FFFFFF} 8 & \cellcolor[HTML]{FFFFFF} 3.2 & \cellcolor[HTML]{FFFFFF} 3 & \cellcolor[HTML]{A5D9B3} 1.8 & \cellcolor[HTML]{72C488} 0.59 & \cellcolor[HTML]{FFFFFF} 2.1 & \cellcolor[HTML]{FFFFFF} 8.2 & \cellcolor[HTML]{FFFFFF} 2.8 & \cellcolor[HTML]{FFFFFF} 2.7 & \cellcolor[HTML]{E4F4E8} 1.4 & \cellcolor[HTML]{ACDCB8} 0.36 & \cellcolor[HTML]{FFFFFF} 2.3 & \textbf{-0.09} \\
\bottomrule
\end{tabular}
}
    \label{tab:nee-sup-90}
\end{table}


\end{document}